%% file: lowrank_SPM_arxiv.tex
\documentclass[10pt]{article}
\usepackage{setspace}
\usepackage[margin=1in]{geometry}

\usepackage{url,mathdots,algorithmic}            
\usepackage{booktabs}       
\usepackage{amsfonts}       
\usepackage{microtype}      
\usepackage{amsmath,amsbsy,amstext,amssymb}
\usepackage{color}
\usepackage{algorithm} 
\usepackage{subfigure}
\usepackage{latexsym,graphics,epsf,psfrag,wrapfig,epsfig}
\usepackage{booktabs,xspace,paralist}

\include{chen_macro}

\usepackage[T1]{fontenc}
\usepackage[latin9]{inputenc}

\title{Harnessing Structures in Big Data \\via Guaranteed Low-Rank Matrix Estimation}

\date{}
\author{Yudong Chen and Yuejie Chi
\thanks{Authors are listed alphabetically.}
\thanks{Y. Chen is with the School of Operations Research and Information Engineering, Cornell University, Ithaca, NY, USA (email: yudong.chen@cornell.edu).}
\thanks{Y. Chi is with the Department of Electrical and Computer Engineering, Carnegie Mellon University, Pittsburgh, PA, USA (email: yuejie.chi@cmu.edu).}}

\newcommand{\bLambda}{{\boldsymbol{\Lambda}}}

\newcommand{\cP}{{\mathcal{P}}}

\newcommand{\bSigma}{{\boldsymbol \Sigma}}
\newcommand{\be}{{\boldsymbol e}}

\newcommand{\bu}{{\boldsymbol u}}
\newcommand{\bv}{{\boldsymbol v}}
\newcommand{\bE}{{\boldsymbol E}}
\newcommand{\bM}{{\boldsymbol M}}

\newcommand{\bA}{{\boldsymbol A}}
\newcommand{\bG}{{\boldsymbol G}}
\newcommand{\bQ}{{\boldsymbol Q}}

\newcommand{\ba}{{\boldsymbol a}}

\newcommand{\bw}{{\boldsymbol w}}
\newcommand{\bS}{{\boldsymbol S}}
\newcommand{\bx}{{\boldsymbol x}}
\newcommand{\bU}{{\boldsymbol U}}
\newcommand{\by}{{\boldsymbol y}}

\newcommand{\bg}{{\boldsymbol g}}
\newcommand{\bL}{{\boldsymbol L}}
\newcommand{\bR}{{\boldsymbol R}}
\newcommand{\bX}{{\boldsymbol X}}

\newcommand{\bV}{{\boldsymbol V}}
\newcommand{\bW}{{\boldsymbol W}}
\newcommand{\bB}{{\boldsymbol B}}

\newcommand{\bY}{{\boldsymbol Y}}
\newcommand{\bP}{{\boldsymbol P}}

\newcommand{\bZ}{{\boldsymbol Z}}

\newcommand{\argmin}{\mathop{\rm argmin}}

\newcommand{\diag}{\mathop{\rm diag}}

\newtheorem{theorem}{\textbf{Theorem}}\newtheorem{definition}{\textbf{Definition}}\newtheorem{remark}{\textbf{Remark}}

\begin{document}

\maketitle
\begin{abstract}

Low-rank modeling plays a pivotal role in signal processing and machine learning, with applications ranging from collaborative filtering, video surveillance, medical imaging, to dimensionality reduction and adaptive filtering. Many modern high-dimensional data and interactions thereof can be modeled as lying approximately in a low-dimensional subspace or manifold, possibly with additional structures, and its proper exploitations lead to significant reduction of costs in sensing, computation and storage. In recent years, there is a plethora of progress in understanding how to exploit low-rank structures using computationally efficient procedures in a provable manner, including both convex and nonconvex approaches. On one side, convex relaxations such as nuclear norm minimization often lead to statistically optimal procedures for estimating low-rank matrices, where first-order methods are developed to address the computational challenges; on the other side, there is emerging evidence that properly designed nonconvex procedures, such as projected gradient descent, often provide globally optimal solutions with a much lower computational cost in many problems. This survey article will provide a unified overview of these recent advances on low-rank matrix estimation from incomplete measurements. Attention is paid to rigorous characterization of the performance of these algorithms, and to problems where the low-rank matrix have additional structural properties that require new algorithmic designs and theoretical analysis. 

\end{abstract}



\section{Introduction}
 
The ubiquity of advanced sensing and imaging technologies produce vast amounts of data at an unprecedented rate. A fundamental goal of signal processing is to extract, and possibly track the evolution of, the relevant structural information faithfully from such high-dimensional data, ideally with a minimal amount of computation, storage and human intervention. To overcome the curse of dimensionality, it is important to exploit the fact that real-world data often possess some {\em low-dimensional geometric structures}. In particular, such structures allow for a succinct description of the data by a number of parameters much smaller than the ambient dimension. One popular postulate of low-dimensional structures is {\em sparsity}, that is, a signal can be represented using a few nonzero coefficients in a proper domain. For instance, a natural image often has a sparse representation in the wavelet domain. The field of {\em compressed sensing} \cite{Candes2006,Donoho2006b} has made tremendous progress in capitalizing on the sparsity structures, particularly in solving under-determined linear systems arising from {\em sample-starved} applications such as medical imaging, spectrum sensing and network monitoring. In these applications, compressed sensing techniques allow for faithful estimation of the signal of interest from a number of measurements that is proportional to the sparsity level --- much fewer than is required by traditional techniques. The power of compressed sensing has made it a disruptive technology in many applications such as magnetic resonance imaging (MRI): a Cardiac Cine scan can now be performed within 25 seconds with the patients breathing freely. This is in sharp contrast to the previous status quo, where the scan takes up to six minutes and the patients need to hold their breaths several times \cite{sudarski2016free}.

While the sparsity model is powerful, the original framework of compressed sensing mainly focuses on vector-valued signals that admit sparse representations in an \emph{a priori known} domain. However, knowledge of such sparsifying domains is not always available, thus limiting its applications. Fortunately, one can resort to a more general notion of sparsity that is more versatile when handling matrix-valued signals --- or an ensemble of vector-valued signals --- without the need of specifying a sparsifying basis. In this paper, we will review this powerful generalization of sparsity, termed the low-rank model, which captures a much broader class of low-dimensional structures. Roughly speaking, this model postulates that the matrix-valued signal is approximately low-rank. If we view each column of the matrix as a data vector, then this is equivalent to saying that the data approximately lies in a low-dimensional but unknown subspace. Historically, the exploitation of low-rank structures may begin even earlier than that of sparsity. In particular, the low-rank assumption is what underlies classical Principal Component Analysis (PCA) \cite{jolliffe1986pca}, which builds on the observation that real-world data has most of its variance in the first few top principal components. Such low-rank structures may arise due to various physical reasons and engineering designs. In face recognition, face images are found to trace out a $ 9 $-dimensional subspace if they are approximately convex and reflect light according to Lambert's law \cite{basri2003lambertian}. In radar and sonar signal processing, the signals reside approximately in a low-dimensional subspace due to transmitting using a small set of waveforms to construct certain beam patterns. Low-rank structures also arise from modeling interactions between different objects. For example,  in clustering or embedding, the pairwise interactions between objects can often be expressed as a low-rank matrix \cite{kulis2007fast}.

Given the collected data, the key problem is to infer the hidden low-dimensional subspace that captures most of the information relevant for subsequent tasks such as detection, clustering, and parameter estimation. Traditional methods such as Singular Value Decomposition (SVD) for finding principal subspaces typically require the data to be fully observed. However, modern data applications often involve estimation problems with a number of measurements that is much smaller than the ambient dimension, a regime similar to the setting of compressed sensing. We refer to this problem as \emph{low-rank matrix estimation}, emphasizing the fact that one only has under-sampled measurements or partial observations. Examples of such problems are abundant. In recommendation systems, the goal is to estimate the missing ratings given a small number of observed ones. In sensor networks, an important problem is to infer the locations of the sensors from pairwise distance measures, which are available only for sensors within a certain radius of each other. In wideband spectrum sensing, to reduce the sampling rate, a popular approach is to estimate the signal subspace and bearing parameters by randomly sub-sampling the outputs of the array. 

In these applications, it is desirable to develop low-rank matrix estimation algorithms that are both \emph{statistically efficient} --- achieving low estimation errors with a minimal amount of (noisy) measurements --- and \emph{computationally efficient} --- having low running time and storage cost. A particular focus of this paper is on algorithms that come with \emph{provable} guarantees for their statistical and computation efficiency. The search for such algorithms is in part motivated by the remarkable success story of compressed sensing, for which many provable methods have been developed for sparse models. Handling the more general low-rank structures poses a new set of challenges as well as opportunities. The study of low-rank matrix estimation has attracted the attention of many researchers from diverse communities including signal processing, machine learning, statistics, mathematical programming and computer science \cite{keshavan2010matrix}\nocite{candes2010power,Negahban2012restricted,ExactMC09}--\cite{Gross2011recovering}.  As we elaborate below, this enterprise has been much fruitful, resulting in many powerful algorithms, novel analytical techniques, and deep theoretical insights.

 This survey article is complementary to the nice overview article on low-rank matrix recovery by Davenport and Romberg \cite{davenport2016overview}, with different focuses. In particular, by focusing on recent algorithmic advancements with computational and statistical guarantees, we highlight the effectiveness of first-order methods in both convex and nonconvex optimization. We also put specific emphasis on a unique set of applications involving structured matrix completion.

\subsection{Paper Organizations}
The rest of this paper is organized as follows. Section~\ref{sec_models} motivates low-rank models from the perspectives of modeling data correlations and lifting vector-valued problems. Section~\ref{sec_setup} describes the basic mathematical setup of the low-rank estimation problem. Section~\ref{sec_convex} discusses the theory and algorithms for low-rank matrix estimation via convex optimization. Section~\ref{sec_nonconvex} discusses the theory and algorithms for low-rank matrix estimation via nonconvex optimization. Section~\ref{sec_structured} discusses structured matrix completion, where the low-rank matrices have additional structural constraints, using several concrete examples. Numerical examples on a real-world recommendation dataset are showcased in Section~\ref{sec_simulations}. The paper is concluded in Section~\ref{sec_conclusions}.

\subsection{Notations} \label{sec:notations}

Throughout this paper, we use boldface capital letters such as $ \boldsymbol{A} $ to denote matrices, with $ \boldsymbol{A}^T $ being its transpose and $ A_{ij} $ being its $ (i,j) $-th entry. Similarly, we use boldface lower-case letters such as $\boldsymbol{a}$ to denote vectors, with $ \boldsymbol{a}^* $ being its conjugate transpose and $ a_i $ being its $ i $-th entry. The expectation is denoted by $\mathbb{E}$.  In addition, $\|\boldsymbol{A}\|$, $\|\boldsymbol{A}\|_{\mathrm{F}}$, $\|\boldsymbol{A}\|_{2,\infty}$, $\mbox{Tr}(\boldsymbol{A})$, and $\|\boldsymbol{A}\|_{*}$, stand for the spectral norm (i.e.~the largest singular value), the Frobenius norm, the $\ell_2/\ell_\infty$ norm (i.e.~the largest $\ell_2$ norm of the rows), the trace, and the nuclear norm (i.e.~the sum of singular values) of the matrix $\boldsymbol{A}$. For two matrices $ \bA $ and $ \bB $ of the same size, $ \langle \bA, \bB \rangle \triangleq \mbox{Tr}(\bA^\top \bB)$ denotes their trace inner product.  The notation $\mbox{diag}[\boldsymbol{c}]$ denotes a diagonal matrix whose diagonal entries are given by the vector $\boldsymbol{c}$. We use $\boldsymbol{e}_i$ to denote the $i$-th standard basis vector of $\mathbb{R}^n$, for each $i = 1,2,\ldots, n$.

\section{The Ubiquity of Low-Rank Models} \label{sec_models}

In this section, we elucidate the motivations for studying low-rank modeling and low-rank matrix estimation problems. We start with a classical viewpoint and justify the low-rank priors of a data matrix from the perspective of bias-variance trade-offs for modeling correlations in data observations. We next argue that low-rank structures arise from a powerful reformulation of quadratic optimization problems by lifting them into a matrix space. Last but not least, we provide a list of other sources of low-rank structures in a wide range of science and engineering problems.

\subsection{Correlation-Aware Modeling of Data Matrices} 

We first motivate the use of low-rank models as a general principle for bias-variance trade-off in signal estimation and processing given noisy data, which is a classical viewpoint articulated by Scharf and Tufts in \cite{scharf1987rank}. Consider a stationary signal $\bx\in\mathbb{R}^n$ with covariance matrix $\bSigma = \mathbb{E}[\bx\bx^T] \in\mathbb{R}^{n\times n}$. 
Suppose that the eigenvalue decomposition of $\bSigma$ is given by $\bSigma =\bU \bLambda \bU^T$, where $\bU=[\bu_1,\cdots, \bu_n]\in\mathbb{R}^{n\times n}$ are the eigenvectors, and $\bLambda=\mbox{diag}[\lambda_1,\ldots,\lambda_n]$ are the eigenvalues arranged as a non-increasing sequence. Define the matrix $\bP_r  \triangleq \bU_r \bU_r^T$,
where $\bU_r =[\bu_1,\ldots,\bu_r] \in \real^{n\times r}$ is the matrix whose columns are the eigenvectors  corresponding to the $r$ largest eigenvalues. One sees that $\bP_r$ is the projection operator onto the subspace spanned by the top-$ r $ eigenvectors $\bu_1,\ldots ,\bu_r$. 

Imagine that we receive a noisy copy of the signal $\boldsymbol{x}$ as
\begin{equation*}
\by = \bx+\bw,
\end{equation*}
 where the elements of the noise vector $ \bw\in\mathbb{R}^n $ are independent with variance $\sigma^2$. Let us estimate the signal $ \bx $ using a reduced-rank model with rank $r$ by projecting observed data onto the $r$-dimensional principal subspace of $\bSigma$. Such an estimate is given by
\begin{equation*}
\hat{\bx}_r = \bP_r \by = \bP_r\bx + \bP_r \bw.
\end{equation*}
We make the crucial observation that one may decompose the mean squared error of the estimate $\hat{\bx}_r  $ into two terms:
\begin{align}
\frac{1}{n}\mathbb{E} \|\hat{\bx}_r  - \bx \|_2^2 &= \frac{1}{n}\mathbb{E}\|\bx - \bP_r\bx \|_2^2 + \frac{1}{n}\mathbb{E}\|\bP_r \bw \|_2^2.
\end{align}
Here the first term corresponds to the model \emph{bias}, 
\[ b_r^2 = \frac{1}{n}\mathbb{E}\|\bx - \bP_r\bx \|_2^2= \frac{1}{n} \left( \sum_{i=r+1}^n \lambda_i^2 \right),
\]
which arises due to approximating a full-rank signal by a rank-$r$ model. The second term corresponds to the model \emph{variance},
\[v_r^2 = \frac{r}{n}\sigma^2,
\]
which arises due to the presence of noise. 

\begin{figure}[t]
\begin{center}
\hspace{-0.1in}\includegraphics[width=0.46\textwidth]{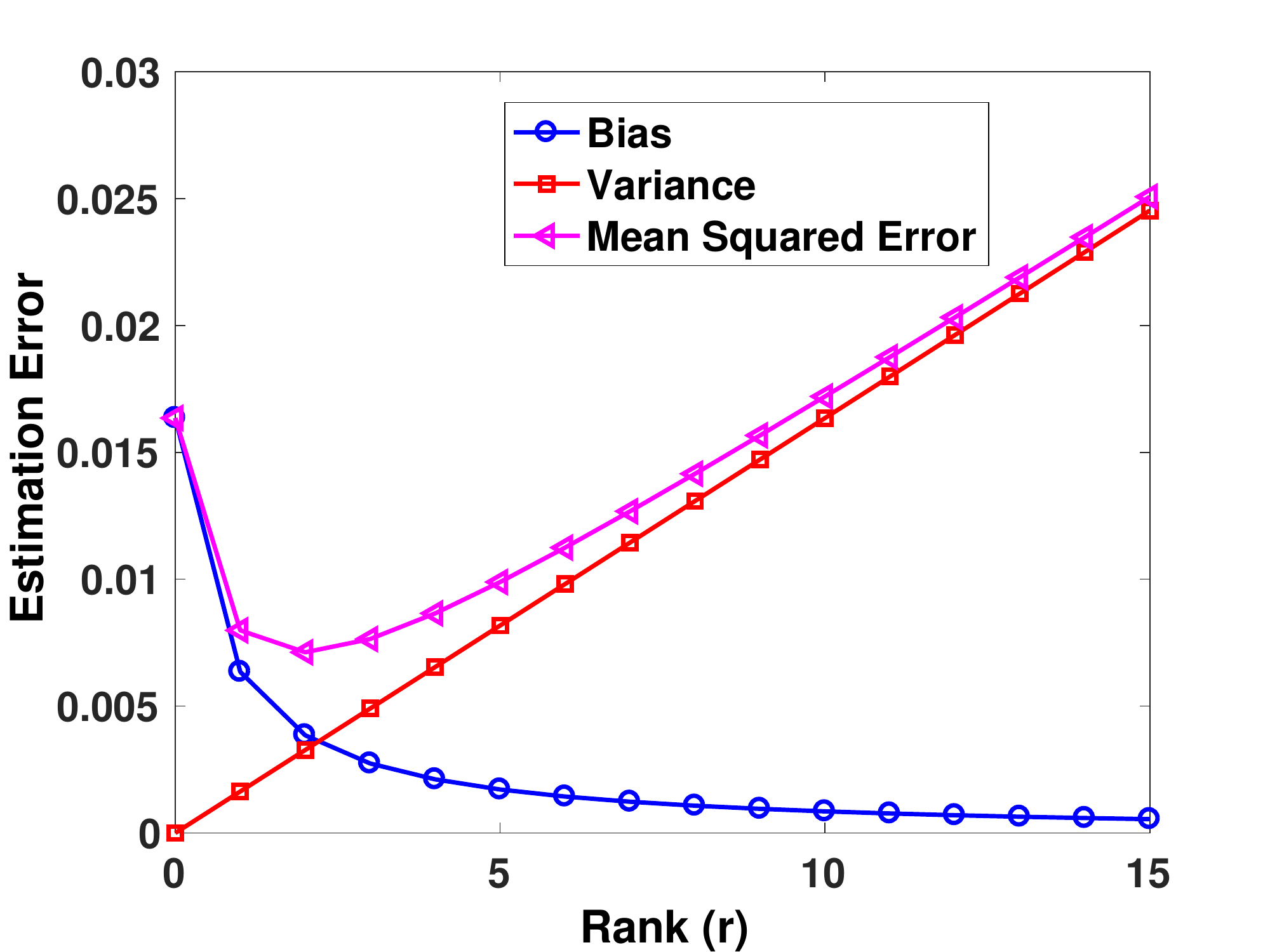}  
\end{center}
\caption{The mean squared error and its decomposition into bias and variance in a signal estimation problem with additive Gaussian noise at a moderate SNR by assuming a low-rank model. This suggests that it is beneficial to apply a reduced-rank model when the data are correlated.}\label{lowrank_model}
\end{figure}

From the above decomposition, we see that as one increases the rank $ r $, the bias of the estimate decreases, whereas the corresponding variance increases. Therefore, the choice of the rank controls the trade-off between the bias and the variance, whose sum constitutes the total estimation error. Importantly, many real-world datasets have a decaying spectrum, which in the above notation means that the eigenvalues of the covariance matrix decrease rapidly. As mentioned, this insight is the foundation of PCA \cite{jolliffe1986pca} and moreover is observed across a wide range of applications including power systems, recommendation systems, Internet traffic and weather data. Consequently, it is beneficial to employ a small rank, so that the variance is controlled, while the bias remains small as long as the residual eigenvalues decreases quickly. With this in mind, we show in Fig.~\ref{lowrank_model} the mean squared error as function of the rank, as well as the decomposition into the bias and the variance; here it is assumed that the spectrum decays at a rate of $\lambda_i = 1/i$, and the signal-to-noise ratio (SNR), defined as $\sum_{i=1}^n\lambda_i^2/(n\sigma^2)$, is equal to $15$dB with $n=100$. It is clear from the figure that employing an appropriate low-rank estimator induces a much lower mean squared error than a full-rank one. In particular, the optimal rank may be much smaller than the ambient dimension $n$ when the spectrum decays fast.

\subsection{Lifting for Quadratic and Bilinear Optimization Problems}
 
Another important source of low-rank structures is solving quadratic/bilinear optimization problems. As an example, consider the \emph{phase retrieval} problem \cite{balan2006signal}, an important routine in X-ray crystallography and optical imaging, where the goal is to recover a vector $\bx$ in $\mathbb{C}^n$ or $\mathbb{R}^n$ given only the \emph{magnitudes} of its linear measurements, that is, 
\begin{equation}
y_l = |\langle \ba_l , \bx\rangle|^2 = \langle \ba_l \ba_l^*, \bx\bx^* \rangle.
\end{equation}
Due to the nonlinear nature of these equations, it is difficult to solve them directly, particularly when the problem size is large. A popular approach for solving such equations is called \emph{lifting}: one rewrites the above equations in terms of the  matrix variable $\bM = \bx \bx^*$, and casts this problem as recovering the rank-one matrix $\bM$ from a set of linear measurements~\cite{candes2012phaselift}. A similar formulation has been used for the blind deconvolution problem; cf.~\cite{ahmed2014blind}.

The lifting approach can be applied to other classes of quadratic equations, whose lifting formulations may lead to low-rank matrices of rank larger than one. For instance, in the problem of sensor network localization \cite{javanmard2013localization}, the goal is to determine the locations of a set of $n$ points/sensors $\{\bx_i\}_{i=1}^n$ lying in an $ r $-dimensional Euclidean space, where $r\ll n$,  given a subset of their pairwise distances. The complete set of pairwise Euclidean distances can be arranged as a matrix $\bE= [E_{ij}]\in\mathbb{R}^{n\times n}$, where $E_{ij} =\|\bx_i-\bx_j\|_2^2$, $1\leq i,j\leq n$. Interestingly, each pairwise distance (that is, an entry of $ \bE $) is in fact a linear function of the rank-$ r $, positive semidefinite (PSD)  matrix $\bM = \bX\bX^T$, where $\bX=[\bx_1,\bx_2,\ldots,\bx_n]^T\in\mathbb{R}^{n\times r}$; more precisely, one has
\begin{equation}
\label{eq:sensor_measurment}
E_{ij} =  \| \bX^T (\be_i-\be_j)\|_2^2 = (\be_i-\be_j)^T \bM (\be_i-\be_j).
\end{equation}
Therefore, the problem of determining the locations $\bX$ of the sensors is equivalent to recovering the low-rank lifted matrix $ \bM $ from a set of linear measurements in the form of~\eqref{eq:sensor_measurment}; see \cite{javanmard2013localization} for a more detailed treatment of this powerful reformulation.

\subsection{Other Sources of Low-rank Structures}
There're many potential sources of low-rank structures. Below we provide a few further examples drawn from different science and engineering domains:
\begin{itemize}
\item In system identification and time series analysis, finding the minimum-order linear time-invariant system is equivalent to minimizing the rank of Hankel structured matrices~\cite{Fazel2003Hankel} (cf.\ Section~\ref{sec:hankel}). 

\item In recommendation systems~\cite{bennett2007netflix}, the matrix of user ratings for a set of items is often approximately low-rank, as user preferences typically depend on a small number of underlying factors and hence their ratings correlate with each other.
\item The background of a video usually changes slowly from frame to frame, hence stacking the frames as columns lead to an approximately low-rank matrix~\cite{candes2009robustPCA}. Similar low-rank structures arise from the smoothness properties of other visual and physical objects~\cite{basri2003lambertian}.
\item In quantum state tomography, the density matrix of a pure or nearly pure quantum state is approximately low-rank, which can be exploited in the problem of state  reconstruction from a small number of Pauli measurements~\cite{gross2010quantum}.
\item In a sparse graphical model with latent variables, one can show, using the Schur complement, that the inverse marginal covariance matrix of the observed variables can be approximated by a matrix with rank equal to the number of latent variables~\cite{chandrasekaran2012latent}.
\item Matrices with certain monotonicity properties can be well-approximated by a matrix with rank much smaller than the ambient dimension. Such matrices arise, for example, when measuring the pairwise comparison scores of a set of objects that possess an underlying ordering~\cite{chatterjee2014universal}.
\item The pairwise affinity matrix of a set of objects is often approximately low-rank due to the presence of clustering/community structures \cite{chen2012sparseclustering} (cf.\ Section~\ref{sec:clustering}).
\end{itemize}
The list continues for much longer. The ubiquity of these structures, either as a physical property or as an engineering choice, is what makes low-rank models useful, and motivates the extensive study of the low-rank matrix estimation problem.

\section{Low-Rank Matrix Estimation from Incomplete Observations}
\label{sec_setup}

In this section, we formally define the problem of {\em low-rank matrix estimation}, that is,  recovery of a low-rank matrix from a number of measurements much smaller than the dimension of the matrix. Let $\boldsymbol{X}\in\mathbb{R}^{n_1\times n_2}$ be the matrix-valued signal of interest.\footnote{Our discussions can be extended complex-valued matrices straightforwardly.}
Denote the SVD of $ \bX $ by
$$\bX=\bU\bSigma\bV^T = \sum_{i=1}^{\min\{n_1,n_2\}} \sigma_i\bu_i\bv_i^T,$$ 
where the singular values $\sigma_1\geq\sigma_2\geq\cdots$ are organized in an non-increasing order. The best rank-$r$ approximation of $ \bX $ is defined as
\begin{equation*}
\bX_r  \triangleq \argmin_{\mathrm{rank}(\bG)\le r} \left\|\bX - \bG\right\|_{\mathrm{F}}.
\end{equation*}
By the Eckart-Young theorem, the optimal approximation $\bX_r$ is given by
\begin{equation}\label{close_form}
\bX_r  = \sum_{i=1}^r \sigma_i \bu_i\bv_i^T.
\end{equation}
Correspondingly, the rank-$r$ approximation error is given by $\|\bX-\bX_r\|_{\mathrm{F}}$, and we say that the matrix $\bX$ is approximately low-rank if its rank-$r$ approximation error is small for some $r\ll \min\{n_1,n_2\}$.

As mentioned, in many modern applications, one does not directly observe $\bX$, but rather is given an under-determined set of indirect noisy measurements of it. Here we assume that one has access to a set of linear measurements in the form
\begin{equation}
y_l = \langle \bA_l , \bX  \rangle + w_l , \quad l=1,\ldots, m,
\end{equation}
where $\bA_l \in\mathbb{R}^{n_1\times n_2}$ is the $l$-th measurement matrix, and $w_l\in\mathbb{R}$ is a noise term. We may rewrite these equations more compactly in a matrix form as
\begin{equation}\label{observe_model}
\by  = \mathcal{A}(\bX) + \bw,
\end{equation}
where $\mathcal{A}: \mathbb{R}^{n_1\times n_2} \to \mathbb{R}^m$ is the linear measurement operator defined by $ [\mathcal{A}(\bX)]_l = \langle \bA_l, \bX \rangle $, and $\bw = [w_1, \ldots ,w_m]^T \in\mathbb{R}^m$ is the vector of noise terms. Denote by $\mathcal{A}^*$ the conjugate operator of $\mathcal{A}$, where $\mathcal{A}^*(\by)=\sum_{l=1}^m y_l \boldsymbol{A}_l$.

As we are primarily concerned with estimating $\bX$ from $m\ll n_1n_2$ measurements, direct approximation via SVD and the Eckart-Young theorem are impossible. Instead, we need to develop alternative methods to find an (approximate) low-rank solution that best fits the set of noisy under-determined linear equations~\eqref{observe_model}. We  further categorize the low-rank matrix estimation problem into two main types based on the structure of the measurement operator:
\begin{itemize}
\item {\em Low-Rank Matrix Sensing}, one observes linear combinations of the entries of $\bX$, where each measurement matrix $\bA_l$ defining the linear combinations is typically dense. 
\item {\em Low-Rank Matrix Completion}, one directly observes a subset of the entries of $\bX$, and aims to interpolate the missing entries. In this case, each $\bA_l$ is a sparse matrix with a single entry equal to $1$ at the corresponding observed index. 
\end{itemize}

For matrix completion, it is convenient to write the measurements in a matrix form as
\begin{equation}
\bY = \mathcal{P}_{\Omega}(\bX)+\bW,
\end{equation}
where $\Omega \subset \{1,2,\ldots, n_1\} \times \{ 1,2,\ldots,n_2\}$ is  the collection of indices of the observed entries, $ \mathcal{P}_{\Omega}: \mathbb{R}^{n_1\times n_2} \to \mathbb{R}^{n_1\times n_2} $ is the entry-wise partial observation operator defined by
\[
[\mathcal{P}_{\Omega}(\bX) ]_{ij} = 
\begin{cases}
X_{ij}, & (i,j)\in \Omega, \\
0, & \mbox{otherwise,} 
\end{cases} 
\]
and $\bW \in \mathbb{R}^{n_1\times n_2} $ is the noise matrix supported on $\Omega$. With this notation, matrix completion is the problem of (approximately) recovering $ \bX $ given $ \bY $ and $ \Omega $.

\section{Theory and Algorithms for Low-Rank Matrix Estimation via Convex Optimization} \label{sec_convex}
 
The development of efficient algorithms for low-rank estimation owes much of its  inspiration to the success of compressed sensing \cite{Candes2006,Donoho2006b}. There, the convex relaxation approach based on  $\ell_1$-minimization is widely used for recovering sparse signals. For low-rank problems, the role of the $ \ell_1 $ norm is replaced by its matrix counterpart, namely the  nuclear norm (also known as the trace norm), which is a convex surrogate for the rank. This idea gives rise to convex optimization approaches for low-rank estimation based on nuclear norm minimization, an approach put forth by Fazel et al.\  in the seminal work~\cite{recht2010guaranteed}. This approach has since been extensively developed and expanded, which remains the most mature and well-understood method (though not the only one) for estimating low-rank matrices. In this section we provide a survey of this algorithmic approach and the associated theoretical results.

\subsection{Convex Relaxation via Nuclear Norm minimization}

We begin by deriving the nuclear norm minimization algorithm as a convex relaxation for rank minimization. Recall our linear measurement model in~\eqref{observe_model}, to which we seek a low-rank solution. A natural approach is to find the matrix with the minimum rank that is consistent with these measurement, which can be formulated as an optimization problem:
\begin{equation}\label{rank_min}
\min_{\bX\in\mathbb{R}^{n_1\times n_2}} ~\mbox{rank}(\bX) \quad \mbox{subject to}\quad  \by = \mathcal{A}(\bX) .
\end{equation}
The rank, however, is a non-convex function of $ \bX $, and rank minimization~\eqref{rank_min} is known to be NP-hard in general. To develop a tractable formulation, one observes that the rank of $\bX$ is equal to the number of its nonzero singular values. Therefore, analogously to using the $ \ell_1 $ norm as a convex surrogate of sparsity, we may replace the rank of $ \bX $ by the sum of its singular values, a quantity known as the nuclear norm:
\begin{equation*}
\|\bX\|_* \triangleq \sum_{i=1}^{\min\{n_1,n_2\}} \sigma_i.
\end{equation*}
Then, instead of solving \eqref{rank_min} directly, one solves for a matrix that minimizes the nuclear norm:
\begin{equation}\label{nuc_norm_min}
\hat{\bX}=\argmin_{\bX\in\mathbb{R}^{n_1\times n_2}} ~\| \bX\|_* \quad \mbox{s.t.}\quad  \by = \mathcal{A}(\bX).
\end{equation}
In the case where the measurements $\by$ are noisy, one seeks a matrix with a small nuclear norm that is approximately consistent with the measurements, which can be formulated either as a regularized optimization problem:
\begin{equation}\label{nuc_norm_noisy}
\hat{\bX} = \argmin_{\bX\in\mathbb{R}^{n_1\times n_2}} ~ \frac{1}{2} \|\by - \mathcal{A}(\bX)\|_2^2 + \tau \|\bX\|_*, 
\end{equation}  
or as a constrained optimization problem:
\begin{equation}\label{nuc_norm_constrained}
\hat{\bX} = \argmin_{\bX\in\mathbb{R}^{n_1\times n_2}} ~ \| \by -\mathcal{A}(\bX)\|_2^2 \quad \mbox{subject to}\;\; \|\bX\|_*\leq \gamma,
\end{equation}
where $\tau$ and $\gamma$ are tuning parameters. Note that the nuclear norm can be represented using the solution to a semidefinite program~\cite{recht2010guaranteed}, 
\begin{align}
\|\bX\|_* =  \min_{\bW_1 , \bW_2 } ~ \frac{1}{2}\left(\mbox{Tr}(\boldsymbol{W}_1)+\mbox{Tr}(\boldsymbol{W}_2)\right)  \label{sdp_trace}\\
\mbox{subject to} \quad \begin{bmatrix}
\boldsymbol{W}_1 & \bX\\
\bX^T & \boldsymbol{W}_2 \end{bmatrix} \succeq 0. \nonumber
\end{align}
Consequently, the optimization problems~\eqref{nuc_norm_min}--\eqref{nuc_norm_constrained} are convex, semidefinite programs.

\subsection{Guarantees for Matrix Sensing via RIP}

For there to be any hope of recovering $ \bX $ from the output of the sensing process~\eqref{observe_model}, the sensing operator $ \mathcal{A} $ needs to possess certain desirable properties so that it can distinguish different low-rank matrices. One such property is called the \emph{restricted isometry property} (RIP). RIP stipulates that $\mathcal{A}$, viewed as a mapping to a lower-dimensional space, preserves the Euclidean distances between low-rank matrices, Below we give a general notion of RIP, where the distances after mapping may be measured in different norms:
\begin{definition}[Restricted Isometry Property] The operator $\mathcal{A}$ is said to satisfy the RIP-$\ell_2/\ell_p$ property of rank $r$ if for all matrices $\boldsymbol{\Phi}$ of rank at most $r$, there holds the inequality
\begin{equation*}
(1-\text{\underbar{$\delta$}}_{r})\left\Vert \boldsymbol{\Phi} \right\Vert _{\mathrm{F}}\leq  \left\Vert \mathcal{A}\left(\boldsymbol{\Phi} \right)\right\Vert _{p}\leq(1+ \bar{\delta}_{r})\left\Vert \boldsymbol{\Phi} \right\Vert _{\mathrm{F}}, \label{eq:ApproximateRIP}
\end{equation*}
where $ \text{\underbar{$\delta$}}_{r} $ and $ \bar{\delta}_{r} $ are some universal constants satisfying  $0<1-\text{\underbar{$\delta$}}_{r}<1<1+\bar{\delta}_{r}$.
\end{definition}

This definition is reminiscent of a similar notion with the same name used in the sparse signal recovery literature that is imposed on sparse vectors~\cite{Candes2006}. Certifying whether RIP holds for a given operator is known to be NP-hard~\cite{bandeira2013certifying}. Nevertheless, it turns out that a ``generic'' sensing operator, drawn from certain random distributions, satisfies RIP with high probability. For example:
\begin{itemize}
\item If the measurement matrix $\boldsymbol{A}_l$ has i.i.d.\ Gaussian entries $\mathcal{N}(0,1/m)$, then $\mathcal{A}$ satisfies RIP-$\ell_2/\ell_2$ with high probability as long as $m\geq c (n_1+n_2)r$ for some large enough constant  $c>0$~\cite{recht2010guaranteed,candes2011tight}. 
\item If the measurement matrix $\boldsymbol{A}_l=\boldsymbol{a}_l\boldsymbol{b}_l^T$ is rank-one with $\boldsymbol{a}_l$, $\boldsymbol{b}_l$ composed with i.i.d.\ Gaussian entries $\mathcal{N}(0,1/m)$, then $\mathcal{A}$ satisfies RIP-$\ell_2/\ell_1$ with high probability as long as $m\geq c (n_1+n_2)r$ for some large enough constant $c>0$~\cite{chen2015exact,cai2015rop}.
\end{itemize}
 
When RIP-$\ell_2/\ell_p$ holds, the nuclear norm minimization approach guarantees exact and stable recovery of the low-rank matrix in both noise-free and noisy cases, as shown in the following theorem adapted from \cite{candes2011tight,chen2015exact,cai2015rop}.
\begin{theorem}\label{thm:ApproxLR} Suppose that the noise satisfies $\left\Vert \bw \right\Vert _{p}\leq\epsilon$. If $\mathcal{A}$ satisfies RIP-$\ell_2/\ell_p$,\footnote{When $p=2$, we require $\bar{\delta}_{4r}<0.1892$, $\text{\underbar{$\delta$}}_{4r}<0.2346$ \cite{candes2011tight}; when $p=1$, we require there exists a universal constant $k\geq 2$ such that $(1+ \bar{\delta}_{kr})/(1-\text{\underbar{$\delta$}}_{kr})<\sqrt{k}$ \cite{chen2015exact,cai2015rop}. Both requirements can be met with a sample complexity of $\mathcal{O}((n_1+n_2)r)$.} then the solution to the nuclear norm minimization algorithms~\eqref{nuc_norm_min}--\eqref{nuc_norm_constrained} (with appropriate values for the tuning parameters) satisfies the error bound
\begin{equation}
\Vert\hat{\bX}-\bX \Vert_{\mathrm{F}}\leq C_{1}\frac{\left\Vert \bX -\bX_r \right\Vert _{*}}{\sqrt{r}}+C_{2} \epsilon \label{eq:ApproxLR}
\end{equation}
simultaneously for all $\bX\in\mathbb{R}^{n_1\times n_2}$, where $C_{1}$ and $C_{2}$
are positive numerical constants. \end{theorem}
 
For studying the performance of nuclear norm minimization via other notions such as the null space property, we refer interested readers to \cite{recht2011null}.

\subsection{Guarantees for Matrix Completion via Incoherence}

In the case of matrix completion, an additional complication arises: it is impossible to recover a low-rank matrix that is also sparse. In particular, when one only samples a small subset of the entries of $ \bX $,  it is very likely that most, if not all, of the nonzero entries of $ \bX $ are missed. This means that the sensing operator $ \mathcal{A} = \ProjObs $ used for matrix completion cannot satisfy the RIP. Therefore, for the problem to be well-posed, we need to restrict attention to low-rank matrices whose mass does not concentrate on a few entries. This property can be formalized by the notion of \emph{incoherence}, which measures the alignment between the column/row spaces of the low-rank matrix with the standard basis vectors:
\begin{definition}[Incoherence] For a matrix $\bU\in\mathbb{R}^{n\times r}$ with orthonormal columns, let $\bP_{\bU}$ be the orthogonal projection onto the column space of $\bU$. The incoherence parameter of $\bU$ is defined as
\begin{equation}\label{coherence}
\mu(\bU) = \frac{n}{r}\max_{1\leq i\leq n}\|\bP_{\bU}\be_i\|_2^2.
\end{equation}
For a matrix with the SVD $\bX=\bU\bSigma\bV^T$, the incoherence parameter of $\bX$ is defined as  
$$\mu_0 = \max\{\mu(\bU),\mu(\bV)\} .$$ 
\end{definition}

It is easy to see that the incoherence parameter satisfies the bound $1\leq\mu(\bU)\leq n/r$. With a smaller $ \mu(\bU) $, the column space of $ \bU $ is more spread out over its coordinates. For a matrix $\bX$, its incoherence parameters $ \mu_0 $ is determined by its the singular vectors and is independent of its singular values. In the noiseless setting, nuclear norm minimization can perfectly recover an incoherent low-rank matrix as soon as the number of measurements is slightly larger than the degrees of freedom of the matrix. Such recovery guarantees were proved and refined in a series of work in \cite{candes2010power,ExactMC09,Gross2011recovering,chen2015incoherence,Recht2009SimplerMC,KeshavanMontanariOh2010}. The theorem below is adapted from \cite{chen2015incoherence}, which is state-of-the-art.
 
\begin{theorem}\cite{chen2015incoherence}\label{mc_theorem} Suppose that each entry of $ \bX $ is observed independently with probability $p\in(0,1)$. If  $ p $ satisfies 
$$ p \geq C  \mu_0 \frac{ r\log^2(n_1+n_2)}{(n_1+n_2)}, $$
for some constant $C$, then with high probability, the nuclear norm minimization algorithm~\eqref{nuc_norm_min} exactly recovers $\bX$ as the unique optimal solution.
\end{theorem}

By a coupon-collecting argument \cite{candes2010power}, one can in fact show that it is impossible to recover the matrix with less than $(n_1+n_2)r\log (n_1+n_2)$ measurements using any algorithm. Therefore, Theorem~\ref{mc_theorem} shows that nuclear norm minimization is near-optimal in terms of sample complexity --- off by only a logarithmic factor --- a remarkable fact considering that we are using a convex relaxation of the rank.

In the noisy case, one can study the performance of nuclear norm minimization in terms of its recovery error, which is done for example in \cite{Negahban2012restricted,koltchinskii2011nuclear}. Here we state one such  performance guarantee taken from \cite{Negahban2012restricted}. Let us assume that the entries of the noise $\bw$ are independent with variance that scaled as $\nu/\sqrt{n_1n_2}$. Then, by solving the regularized nuclear norm minimization problem~\eqref{nuc_norm_noisy} with parameter $\tau = 4\nu \sqrt{\frac{(n_1+n_2)
\log (n_1+n_2)}{m}}$, we obtain a solution $ \hat{\bX} $ satisfying
\[ \|\hat{\bX} - \bX\|_{\mathrm{F}}^2 \leq \nu^2 \frac{(n_1+n_2)r \log (n_1+n_2)}{m} \]
with high probability in the moderate to low SNR regime.

\subsection{First-Order Algorithms for Nuclear Norm Minimization}

In principle, it is possible to solve the nuclear norm minimization problems in~\eqref{nuc_norm_min}--\eqref{nuc_norm_constrained} to high numerical accuracy using off-the-shelf semidefinite programming solvers (such as SDPT3 \cite{toh1999sdpt3}). However, these solvers, typically based on interior-point methods, can be extremely slow when the size of the matrix is large. For example, SDPT3 can only handle matrices with dimensions no larger than a few thousands due to memory requirements. This computational issue motivates the development of fast alternatives that can handle significantly larger problems. First-order algorithms become an appealing candidate due to their low per-iteration cost, as well as the flexibility to incorporate the specific structures of the semidefinite programs that arise in low-rank matrix estimation. There is a long and still growing list of such algorithms, including singular value thresholding~\cite{cai2010singular}, accelerated proximal gradient descent~\cite{toh2010accelerated}, which is variant of FISTA for matrix completion \cite{beck2009fast}, Augmented Lagrangian Multiplier methods~\cite{Lin2009_ALM}, Frank-Wolfe \cite{jaggi2013revisiting,freund2017extended}, CoGENT \cite{rao2015forward}, and ADCG~\cite{boyd2017alternating}, just to name a few. Below, we discuss two representative algorithms: FISTA for solving the regularized problem \eqref{nuc_norm_noisy}, and Frank-Wolfe for solving the constrained problem \eqref{nuc_norm_constrained}. These two algorithms provide the stage for understanding many other algorithms.

An important subroutine in many of the aforementioned algorithms is the Singular Value Thresholding (SVT) operator $ \mathcal{D}_\tau(\cdot) $~\cite{cai2010singular}. Mathematically, $ \mathcal{D}_\tau(\bY) $ is defined as the proximal mapping of $\bY$ with respect to the nuclear norm:
\begin{equation}
\mathcal{D}_{\tau}(\bY)  = \argmin_{\bZ} \frac{1}{2}\|\bY -\bZ\|_{\mathrm{F}}^2 + \tau \|\bY\|_*.
\end{equation}
The SVT operator admits a closed-form expression; in particular, if the SVD of $\bY$ is $\bU\bSigma\bV^T$ with $ \bSigma = \diag[\sigma_1,\sigma_2,\ldots] $, then $\mathcal{D}_{\tau}(\bY)  = \bU\bSigma'\bV^T$, where $ \bSigma' = \diag[\sigma'_1,\sigma'_2,\ldots]$ with
\begin{equation}
\sigma_k' = 
\begin{cases}
\sigma_k - \tau, & \quad \sigma_k\geq \tau, \\
0, &\quad \sigma_k<\tau.
\end{cases} 
\end{equation}
The fact that the SVT operator can be efficiently computed via SVD is leveraged in many of the first-order algorithms.

\begin{algorithm}[t]
\caption{FISTA for Low-Rank Matrix Estimation}

\label{alg:fista}\begin{algorithmic}

\STATE \textbf{{Parameters}}: $T$, $L$, $\tau$ in \eqref{nuc_norm_noisy};

\STATE \textbf{{Initialization}}: Let $\boldsymbol{R}_1=\boldsymbol{0}$, $t_1 =1$;

\STATE \textbf{{Updates}}: \textbf{for} $k=1,2,\ldots, T$
\textbf{do}

\STATE
\begin{enumerate}
\item $\bX_k = \mathcal{D}_{\tau/L}\left(\bR_k -\frac{1}{L} \mathcal{A}^*(\mathcal{A}(\bR_k)-\by)\right)$;
\item $t_{k+1} = \frac{1+\sqrt{1+4t_k^2}}{2}$;
\item $\bR_{k+1} = \bX_k + \left(\frac{t_k-1}{t_{k+1}}\right)(\bX_k - \bX_{k-1})$.
\end{enumerate}

\STATE \textbf{{Output}}: $\bX_T$.
\end{algorithmic} 
\end{algorithm}

The FISTA algorithm for the regularized problem~\eqref{nuc_norm_noisy} is given in Algorithm~\ref{alg:fista}, where $L$ is an upper bound of the Lipschitz constant of $\nabla f(\bX)$, with $f(\bX):=\frac{1}{2} \|\by - \mathcal{A}(\bX)\|_2^2$. FISTA makes use of Nesterov's momentum acceleration to speed up the convergence. If we denote the objective function of \eqref{nuc_norm_noisy} as $g(\bX)$, then to achieve $\epsilon$-accuracy, i.e. $g(\bX_T) - g(\hat{\bX})\leq \epsilon$, we need $T=\mathcal{O}\left(\sqrt{L/\epsilon}\right)$ iterations. In each iteration, only a partial SVD is needed to evaluate the SVT operator. Doing so for large-scale problems may still be too slow and require large memory. In this case, one can make use of modern randomized techniques from numerical linear algebra to further speed up the computation of SVD~\cite{cevher2014convex,halko2011finding}.

The standard Frank-Wolfe method~\cite{jaggi2013revisiting} (also known as conditional gradient descent) for solving the constrained problem \eqref{nuc_norm_constrained} is presented in Algorithm~\ref{alg:frank_wolfe}. Each iteration of algorithm only requires computing a rank-one SVD, which can be done using power iteration or Lanczos methods. Therefore, Frank-Wolfe typically has a much lower computational cost per-iteration than methods based on the SVT operation. However, standard Frank-Wolfe may converge very slowly in practice. To achieve $\epsilon$-accuracy, i.e. $f(\bX_T) - f(\hat{\bX})\leq \epsilon$, Frank-Wolfe requires $T=\mathcal{O}\left(1/\epsilon\right)$ iterations, which can be quite slow.   
Variants of Frank-Wolfe with faster convergence or lower memory footprint have been actively developed recently by exploiting the problem structures. The list, including CoGENT \cite{rao2015forward}, In-Face Extended Frank-Wolf \cite{freund2017extended}, and ADCG \cite{boyd2017alternating}, Block Frank-Wolfe \cite{allen2017linear}, and sketchyCGM \cite{yurtsever2017sketchy}, is still growing.

\begin{algorithm}[t]
\caption{Frank-Wolfe for Low-Rank Matrix Estimation}

\label{alg:frank_wolfe}\begin{algorithmic}

\STATE \textbf{{Input}}: $T$, $\gamma$ in \eqref{nuc_norm_constrained};

\STATE \textbf{{Initialization}}: Let $\boldsymbol{X}_1=\boldsymbol{0}$, 

\STATE \textbf{{Updates}}: \textbf{for} $k=0,1,2,\ldots, T-1$
\textbf{do}

\STATE
\begin{enumerate}
\item $\bS_{k} = \gamma \bu\bv^T$, where $\bu$ and $\bv$ are the left and right top singular vector of $\mathcal{A}^*(\by-\mathcal{A}(\bX_k)))$;
\item $\bX_{k+1}=(1-\frac{2}{k+1})\bX_{k} + \frac{2}{k+1} \bS_{k}$;
\end{enumerate}
\STATE \textbf{{Output}}: $\bX_{T}$. 
\end{algorithmic} 
\end{algorithm}

\section{Provable and Fast Low-Rank Matrix Estimation via Nonconvex Factorization}
\label{sec_nonconvex}
 
As we have seen, the computational concern of solving rank minimization problems is assuaged to some extent by the use of convex relaxation --- the resulting semidefinite programs can be solved in time polynomial in the matrix dimension. However, for large-scale problems where the dimension is on the order of millions, solving these semidefinite programs, even using first-order methods, can still be computationally infeasible due to the fundamental bottleneck of storing and optimizing over a matrix variable. This issue severely limits the applicability of the convex relaxation methods. 

To overcome this difficulty, a recent line of work studies more computationally efficient  methods that are based on nonconvex optimization. These methods work directly with the original nonconvex, rank-constrained optimization problem, which can be generally written as
	\begin{align} \label{eq:low-rank}
	\min_{\PlTheta\in \real^{n_1 \times n_2}} & \;\; \EmpLoss(\PlTheta)  \quad\mbox{subject to} \quad \rank(\PlTheta) \le \rdim, 
	\end{align}
where $ \EmpLoss: \real^{n_1 \times n_2} \to \real$ is a given loss function, which typically is convex in $\bX$. The key idea is to use a reparametrization trick:
by writing a rank-$ \rdim $ matrix in its factorization form $ \PlTheta = \CROSS{\ltheta}{\rtheta}  $, where $ \ltheta \in \real^{n_1 \times \rdim} $ and $ \rtheta \in \real^{n_2 \times \rdim} $, we enforce the low-rank constraint directly, leading to the following equivalent formulation of~\eqref{eq:low-rank}:
\begin{align}
	\min_{\ltheta \in \real^{n_1 \times \rdim},\rtheta \in \real^{n_2 \times \rdim}} & \quad \EmpFLoss(\ltheta, \rtheta) \defn \EmpLoss(\CROSS{\ltheta}{\rtheta}) . \label{eq:nonconvex_prog}
	\end{align}

We refer to this formulation as the \emph{\BM} factorization, after the seminal work~\cite{burer2005LRSDP}.
The optimization problem~\eqref{eq:nonconvex_prog} can then be solved \emph{over the factor variables} $ \ltheta $ and $ \rtheta $.  The low-rank factorization $ \bX = \CROSS{\ltheta}{\rtheta} $ is in general not unique; in fact, any pair $ \widetilde{\ltheta} = \ltheta \bQ  $ and $ \widetilde{\rtheta} = \rtheta \bQ $ with $ \bQ \in \real^{\rdim \times \rdim } $ being an orthonormal matrix also corresponds to the same matrix $ \bX $, since $ \widetilde{\ltheta} \widetilde{\rtheta}^T = \ltheta \bQ \bQ^T \rtheta = \ltheta \rtheta$. 
These pairs are all global optima of the problem under certain conditions to be discussed below. This reformulation brings a significant computational gain: since 
the rank $ \rdim $ is often much smaller than~$\min\{n_1,n_2\}$,  the size of the variables $ (\ltheta, \rtheta)$ is roughly linear in $ (n_1+n_2) $ rather than quadratic, leading to the possibility of designing \emph{linear-time} algorithms that are amenable to problems of very large scale.

Surprisingly, even though the \BM formulation~\eqref{eq:nonconvex_prog} is nonconvex, global optima can sometimes be found (or approximated) efficiently using various iterative procedures; moreover, rigorous guarantees can be derived for the statistical accuracy of the resulting solution. Indeed, several iterative schemes have a computational cost proportional to $(n_1+n_2) \poly(\rdim) $ and the size of the input, at least per iteration, which is typically much lower than $ n_1\times n_2 $.  These results are developed in a still growing line of recent work~\cite{chenwainwright}\nocite{keshavan2010matrix,sun2014mc,tu2016low,zheng2015convergent,zheng2016rectangular,GeLeeMa2016,bhojanapalli2016global,li2016rapid,sun2016geometric,chen2015quadratic,jain2013altMin,hardt2014understanding,haeffele2015global,jain2010svp,jain2014stageSVP,netrapalli2014nonconvexRPCA,candes2014wirtinger,zhang2016median,zhang2016median,ma2017implicit,li2018nonconvex}--\cite{bandeira2016low}, and we devote the rest of this section to presenting the most representative results therein.

This line of work considers three major classes of iterative schemes for solving the \BM formulation~\eqref{eq:nonconvex_prog}:
\begin{itemize}
	\item \textbf{(Projected) gradient descent}~\cite{burer2005LRSDP,chenwainwright,ma2017implicit}: One runs (projected) gradient descent directly on the loss function $ \EmpFLoss(\ltheta, \rtheta)$ with respect to the factor variables $ (\ltheta,\rtheta) $:
	\begin{subequations}\label{eq:grad}
	\begin{align}
	\lthetait{t+1}& = \Proj_{\LCons } \Big[\lthetait{t} - \stepit{t} \nabla_{\ltheta} \EmpFLoss(\lthetait{t}, \rthetait{t}) \Big], \\
	\rthetait{t+1} &= \Proj_{\RCons} \Big[  \rthetait{t}- \stepit{t} \nabla_{\rtheta} \EmpFLoss(\lthetait{t}, \rthetait{t}) \Big],
	\end{align}
	\end{subequations}
	where $ \stepit{t} $ is the step size and $ \Proj_{\LCons} $,  $ \Proj_{\RCons} $  denote the Euclidean projection onto the sets $ \LCons$ and $ \RCons $, which are constraint sets that encode additional structures of the desired low-rank factors.
	
	\item \textbf{Alternating minimization}~\cite{jain2013altMin,hardt2014understanding}: One optimizes the loss function $ f(\ltheta, \rtheta) $ alternatively over one of the factors while fixing the other, which is a convex problem. In particular, each iteration takes the form
	\begin{subequations}\label{eq:AM}
	\begin{align}
	\lthetait{t+1} &= \argmin_{\ltheta \in \mathbb{R}^{n_1\times r}} \EmpFLoss(\ltheta, \rthetait{t}), \\
	\rthetait{t+1} &= \argmin_{\rtheta \in \mathbb{R}^{n_2\times r} } \EmpFLoss(\lthetait{t+1}, \rtheta). 
	\end{align}
	\end{subequations} 
	
	\item \textbf{Singular value projection (SVP)}~\cite{jain2010svp,jain2014stageSVP,netrapalli2014nonconvexRPCA}: One performs a gradient descent step of $ \EmpLoss(\bL\bR^T) $ on the ``full'' $ n_1 \times n_2 $ matrix space, then projects back to the factor space via SVD:
	\begin{align}\label{eq:svp}
	(\lthetait{t+1}\!, \rthetait{t+1})  \!=\! \SVD_\rdim \Big[ \lthetait{t}\boldsymbol{R}^{tT} \!\! - \stepit{t} \nabla \EmpLoss( \lthetait{t} \boldsymbol{R}^{tT} ) \Big] ,
	\end{align}
	where $ \stepit{t} $ is the step size, and $ \SVD_\rdim (\bZ) $ returns the top rank-$ \rdim $ factors of  $ \bZ $, that is, the pair $ (\bU \bSigma ^{1/2}, \bV \bSigma ^{1/2} )  $ assuming that $ \bU\bSigma \bV^T  = \bZ_r= \Proj_r(\bZ) $ is SVD of the best rank-$ \rdim$ approximation of $ \bZ $.

\end{itemize}

Because neither the function $ \EmpFLoss(\ltheta,\rtheta)$ nor the set of low-rank matrices are convex, standard global convergence theory for convex optimization does not apply here. 
The recent breakthrough is based on the realization that convexity is in fact not necessary for the convergence of these iterative schemes; instead, as long as the gradients of function $ \EmpFLoss(\ltheta,\rtheta)$ always point  (approximately) towards the desired solution, the iterates will make progress along the right direction. Note that this property concerns the \emph{geometry} of $ \EmpFLoss(\ltheta,\rtheta)$ itself, and is largely independent of the specific choice of the algorithm \cite{sun2014mc}. Among the above three options, the projected gradient descent approach stands out due to its simple form, cheap per-iteration cost (no SVD or inner optimization is needed) and efficiency with constrained problems. We thus use projected gradient descent as the focal point of our survey. 

Playing a key role here is the use of statistical modeling and probabilistic analysis: we will show that $ \EmpFLoss(\ltheta,\rtheta) $ has the desired geometric properties with high probability under probabilistic generative models of the data, thereby circumventing the worst-case hardness of the low-rank matrix estimation problem and instead focusing on its average-case behavior. Existing results in this direction can be divided into two categories. In the first line of work reviewed in Section~\ref{sec:local}, one shows that iterative algorithms converge to the desired solution rapidly when initialized within a large neighborhood around the ground truth; moreover, a good initial solution can be obtained efficiently by simple procedures (which typically involve computing a partial SVD). The second line of work, reviewed in Section~\ref{sec:global}, concerns the global landscape of the loss function, and aims to show that in spite of the non-convexity of $ \EmpFLoss(\ltheta,\rtheta) $, all of its local minima are in fact close to the desired solution in an appropriate sense whereas all other stationary points (e.g., saddle points) possess a descent direction; these properties guarantee the convergence of iterative algorithms from any initial solution. Both of these two types of results have their own merits and hence are complementary to each other. Below we review the most representative results in each of these two categories for noiseless matrix sensing and matrix completion.  The readers are referred to \cite{chenwainwright,ma2017implicit,ge2017no} for extensions to the noisy case.

\subsection{Convergence Guarantees with Proper Initialization}~\label{sec:local}

For simplicity, we assume that the truth $\bX$ is exactly rank-$r$ and has a bounded condition number $\kappa = \sigma_1/\sigma_r$,  thus effectively hiding the dependence on $ \kappa $. Moreover, to measure the convergence of the algorithms for the purpose of reconstructing $\bX$, we shall consider directly the \emph{reconstruction error} $  \|\lthetait{t}\boldsymbol{R}^{tT} - \bX \|_{\mathrm{F}} $ with respect to $ \bX$.

For matrix sensing, we take the loss function $\LossTil(\ltheta,\rtheta) $ as
\begin{align}\label{eq:loss_sensing}
\LossTil_{\mathcal{A}}(\ltheta,\rtheta) =   \twonorm{ \Xmap(\CROSS{\ltheta}{\rtheta}) - \yout }^2+ \frac{1}{8} \|\ltheta^T\ltheta - \rtheta^T \rtheta\|_{\mathrm{F}}^2,
\end{align}
where the second regularization term encourages $ \ltheta $ and $ \rtheta $ to have the same scale and ensures algorithmic stability. We can perform gradient descent as specified in \eqref{eq:grad} with  $ \LCons = \real^{n_1 \times \rdim} $ and $  \RCons= \real^{n_2 \times \rdim}$ (that is, without additional projections).
 
Similarly, for matrix completion, we use the loss function
\begin{align}\label{eq:obj_proj_grad}
\EmpFLoss_{\Omega}(\ltheta,\rtheta)   = \frac{1}{\pobs} \| \ProjObs(\CROSS{\ltheta}{\rtheta} - \Yout) \|_{\mathrm{F}}^2 + \frac{1}{32} \|\ltheta^T\ltheta - \rtheta^T \rtheta\|_{\mathrm{F}}^2.
\end{align} 
Since we can only hope to recover matrices that satisfy the incoherence property, we perform projected gradient descent by projecting to the constraint set:
\begin{align*}
\LCons  \defn \Big\{\ltheta \in \real^{n_1 \times \rdim} \,
\mid\, \twoinfnorm{\ltheta} \le
\sqrt{\frac{2\inco \rdim}{\usedim_1}}\opnorm{\lthetait{0}} \Big\},
\end{align*}
with $ \RCons $ defined similarly.
Note that $ \LCons $ is convex, and depends on the initial solution
$ \lthetait{0} $.  The projection
$\ \Proj_{\LCons} $  is given by the row-wise ``clipping'' operation
\begin{align*}
[\Proj_{\LCons}(\ltheta)]_{i\cdot} = 
\begin{cases} 
\ltheta_{i\cdot}\;, & \twonorm{\ltheta_{i\cdot}} \le
\sqrt{\frac{2\inco\rdim}{\usedim_1}}
\opnorm{\lthetait{0}},\\ 
\ltheta_{i\cdot}
\sqrt{\frac{2\inco\rdim}{\usedim_1}}
\frac{\opnorm{\lthetait{0}}}{\twonorm{\ltheta_{i\cdot}}} \;, &
\twonorm{\ltheta_{i\cdot}} > \sqrt{\frac{2\inco\rdim}{\usedim_1}}
\opnorm{\lthetait{0}},
\end{cases}
\end{align*}
for $i = 1,2,\ldots,n_1$, where $ \ltheta_{i\cdot} $ is the $ i $-th row of $ \ltheta $; the projection $\Proj_{\RCons} $ is given by a similar formula.
This projection ensures that the iterates of projected gradient
descent~\eqref{eq:grad} remain incoherent.

The following theorems guarantee that if the initial solution $ (\lthetait{0}, \rthetait{0}) $ is reasonably close to the desired solution,  then the iterates converge linearly  to the ground truth (in terms of the reconstruction error), under conditions similar to nuclear norm minimization. Moreover, we can find a provably good initial solution using the so-called spectral method, which involves performing a partial SVD. In particular, we compute $(\lthetait{0}, \rthetait{0}) = \SVD_\rdim [  \Xmap^{*}(\yout) ]$ for matrix sensing, and $ (\lthetait{0}, \rthetait{0}) = \SVD_\rdim [  \pobs^{-1} \ProjObs(\Yout)] $ for matrix completion.
 
\begin{theorem}[Matrix Sensing]\cite{chenwainwright,jain2010svp,tu2016low}
	\label{thm:local_sensing}
	Suppose  that the sensing operator $\Xmap$ satisfies RIP-$\ell_2/\ell_2$ with parameter $ \ripparam{4\rdim}=\max\{\text{\underbar{$\delta$}}_{4r}, \bar{\delta}_{4r}\} \leq  c_1/\rdim $ for some sufficiently small constant $c_1$. Then, the initial solution $(\lthetait{0}, \rthetait{0}) = \SVD_\rdim [ \Xmap^{*}(\yout) ]$ satisfies
	\begin{align}\label{initialization_condition}
		\|\lthetait{0} \boldsymbol{R}^{0T} - \bX \|_{\mathrm{F}} \leq c_0\sigma_\rdim,
	\end{align}
where $c_0$ is some sufficiently small constant. Furthermore, starting from any $(\lthetait{0}, \rthetait{0})$ that satisfies \eqref{initialization_condition}, the gradient descent iterates $\{(\lthetait{t}, \rthetait{t} ) \}_{t=1}^{\infty}$ with an appropriate step size satisfy the bound
	\begin{align*}
		\| \lthetait{t} \boldsymbol{R}^{tT}- \bX \|_{\mathrm{F}}^2  
		& \le  (1-\delta)^{t} 	\| \lthetait{0} \boldsymbol{R}^{0T}- \bX \|_{\mathrm{F}}^2  ,
	\end{align*}
where $0<\delta<1$ is a universal constant.
\end{theorem}

\begin{theorem}[Matrix Completion]\cite{yi2016rpca}\label{thm:initial_partial}
	There exists a positive constant $C $ such that if 
	\begin{equation} \label{eq:condition_thm_initial_partial}
	\pobs \geq C \frac{\inco_0^2 r^2 \log (n_1+n_2)}{(n_1+n_2)}, 
	\end{equation}
	then with probability at least $1 - c (n_1+n_2)^{-1}$ for some positive constant $c$, the initial solution satisfies 
	\begin{equation}\label{eq:initial_basin}
	\| \lthetait{0} \boldsymbol{R}^{0T}- \bX \|_{\mathrm{F}}  \leq c_0\sigma_\rdim.
	\end{equation}
	Furthermore, starting from any $(\lthetait{0}, \rthetait{0})$ that satisfies \eqref{eq:initial_basin}, the projected gradient descent iterates $\{(\lthetait{t},\rthetait{t})\}_{t = 0}^{\infty}$ with an appropriate step size satisfy the bound
	\[
	\| \lthetait{t} \boldsymbol{R}^{tT}- \bX \|_{\mathrm{F}}^2  
	\leq \left(1 - \frac{\delta}{ \inco_0 \rdim }\right)^t  \| \lthetait{0} \boldsymbol{R}^{0T}- \bX \|_{\mathrm{F}}^2 , 
	\]
	where $0<\delta<1$ is a universal constant. 

\end{theorem}

The above theorems guarantees that the gradient descent iterates enjoy geometric convergence to a global optimum when the initial solution is sufficiently close to the ground truth. Comparing with the guarantees for nuclear norm minimization, (projected) gradient descent succeeds under a similar sample complexity condition (up to a polynomial term in $r$ and $\log n$), but the computational cost is significantly lower. To obtain $\varepsilon$-accuracy, meaning that the final estimate $(\widehat{\bL},\widehat{\bR})$ satisfies 
	\begin{equation} \label{eq:final_error_partial}
	\|\widehat{\bL}\widehat{\bR}{}^T - \bX \|_{\mathrm{F}} \leq \varepsilon\cdot\sigma_\rdim,
	\end{equation}
we only need to run a total of $T=\mathcal{O}(\log(1/\varepsilon))$ iterations for matrix sensing, and $T=\mathcal{O}(\inco_0 \rdim \log (1/\varepsilon))$ iterations for matrix completion.

We now discuss the overall computational complexity, and for simplicity we shall assume $ n_1 = n_2 = n $. For matrix sensing, let $T_{\text{0}}$ be the maximum time of multiplying the matrix $\Xit{l}$ with a
vector of compatible dimension. Each gradient step~\eqref{eq:grad} can be performed in time $\order(\numobs\rdim T_{\text{0}} +n\rdim)$. The complexity of computing the initial solution is a bit subtler. To this end, we first note that by standard matrix perturbation bounds, one can show that the matrix $  \Xmap^{*}(\yout) $ and its singular values are sufficiently close to those of $ \bX $ under the condition of Theorem~\ref{thm:local_sensing}; in particular, one has $ \opnorm{\Xmap^{*}(\yout) - \bX} \le c_0 \sigma_r/\sqrt{r}$,  and the $ r $-th and $ (r+1) $-th singular values of $ \Xmap^{*}(\yout) $ are at most $c_0 \sigma_r/\sqrt{r} $ away from the corresponding singular values of $ \bX $, where $ c_0 $ is a small constant. With such properties of $ \Xmap^{*}(\yout) $, one does \emph{not} need to compute the exact \emph{singular values/vectors} of $ \Xmap^{*}(\yout) $ in order to meet the initialization condition~\eqref{initialization_condition}; rather, it suffices to find a \emph{rank-$ r $ approximation} of $ \Xmap^{*}(\yout)  $  with the property $ \opnorm{\boldsymbol{L}^0\boldsymbol{R}^{0T}- \Xmap^{*}(\yout)} \le c_0 \sigma_r/\sqrt{r}  $. This can be done using for example the Randomized SVD procedure in~\cite{halko2011finding}, which takes $\order(\numobs\rdim T_{\text{0}}\log n + n\rdim^{2})$ time to compute; see \cite[Theorem~1.2 and Eq.~(1.11)]{halko2011finding} with $ q=\log n $. Put together, the overall time complexity is $\order(m\rdim T_{\text{0}}\log(n/\varepsilon))$ for achieving $\varepsilon$-accuracy.

For matrix completion, to obtain the initial solution, we can again follow similar arguments as above to show that it suffices to compute a rank-$ r $ approximation of the matrix $  {\pobs}^{-1} \ProjObs(\Yout) $, which is close to $ \bX $ and has a sufficiently large spectral gap. Since $  {\pobs}^{-1} \ProjObs(\Yout) $ is a sparse matrix with support~$\Obs$, computing such an approximation can be done in time $\order(\rdim |\Obs| \log n + nr^2 )$ using the Randomize SVD procedure in~\cite{halko2011finding}. Each step of gradient descent requires computing the gradient and the projection onto $ \LCons$  and $\RCons$. Both of them only involve operations on sparse matrices supported on $ \Obs $ and thin matrices, and can be done in time $\order(\rdim |\Obs| +  n\rdim^2)$. Therefore, projected gradient descent  achieves $ \epsilon $-accuracy with running time $\order(\rdim |\Obs|  \log n \log(1/\varepsilon))$.

\begin{remark}
Via a refined analysis of gradient descent, it is in fact possible to drop the projection step onto $\LCons$ and $\RCons$ in matrix completion without performance loss; cf. \cite{ma2017implicit}. In particular, as long as $\pobs \geq C \frac{\inco_0^3 r^3 \log^3 (n_1+n_2)}{(n_1+n_2)}$ for some constant $C$, gradient descent converges geometrically, and needs $T=\mathcal{O}(  \log (1/\varepsilon))$ iterations to reach $\epsilon$-accuracy for the reconstruction error measured not only in the Frobenius norm $\|\lthetait{t}\boldsymbol{R}^{tT} - \bX \|_{\mathrm{F}}$, but also in the spectral norm $\|\lthetait{t}\boldsymbol{R}^{tT} - \bX \|$ and the entry-wise infinity norm $\|\lthetait{t}\boldsymbol{R}^{tT} - \bX \|_{\infty}$. For the Singular Value Projection algorithm~\eqref{eq:svp}, geometric convergence in entry-wise infinity norm is also established in the work~\cite{ding2018leave} without the need of additional regularization or separate initialization procedure.
\end{remark}

\begin{remark}
In the noisy setting, the algorithms can be applied without change, and the same error bounds hold with an additional term that depends on the noise. For matrix completion \cite{chenwainwright}, this term is $ \frac{\sqrt{\rdim} \opnorm{ \ProjObs( \noisemat)}}{\pobs} $, where $ \ProjObs( \noisemat) $ is the noise matrix supported on the observed indices. This term can be bounded under various noise models. For example, when $ \noisemat $ has i.i.d.\ Gaussian or $ \pm \noisestd $ entries with zero mean and variance $ \noisestd^2 $, then $ \opnorm{ \ProjObs( \noisemat)} \lesssim \noisestd \sqrt{ \pobs (n_1+n_2)} $ with high probability. The resulting error bound is optimal in an information-theoretic sense~\cite{Negahban2012restricted}. See also \cite{ma2017implicit} for the near-optimal error control in the spectral norm and the entry-wise infinity norm.
\end{remark}

\subsection{Global Geometry and Saddle-Point Escaping Algorithms}\label{sec:global}

A very recent line of work studies the \emph{global} geometry of the \BM formulation~\eqref{eq:nonconvex_prog}, as well as its computational implications for algorithms starting at an \emph{arbitrary} initial solution~\cite{GeLeeMa2016,bhojanapalli2016global,bandeira2016low,sun2016geometric}. These results are based on a geometric notion called the \emph{strict saddle property}~\cite{GeLeeMa2016, ge2017no}.
\begin{definition}[Strict Saddle]
	\label{def:strict_saddle}
A function $ g(\bx) $ is said to be $ (\epsilon, \gamma, \zeta) $-strict saddle, if for each $ \bx $ at least one of the following holds:
	\begin{itemize}
		\item $ \twonorm{\nabla g(\bx) } \ge \epsilon >0 $;
		\item $ \lambda_{\min} \left( \nabla^2 g(\bx)\right) \le -\gamma <0 $;
		\item there exists a local minimum $ \bx^{\star}$ such that $ \twonorm{\bx - \bx^{\star}} \le \zeta$.
	\end{itemize}
\end{definition}

\begin{figure}[t]
	\begin{center}
		\hspace{-0.1in}\includegraphics[width=0.4\textwidth, clip, trim = 200 90 410 180  ]{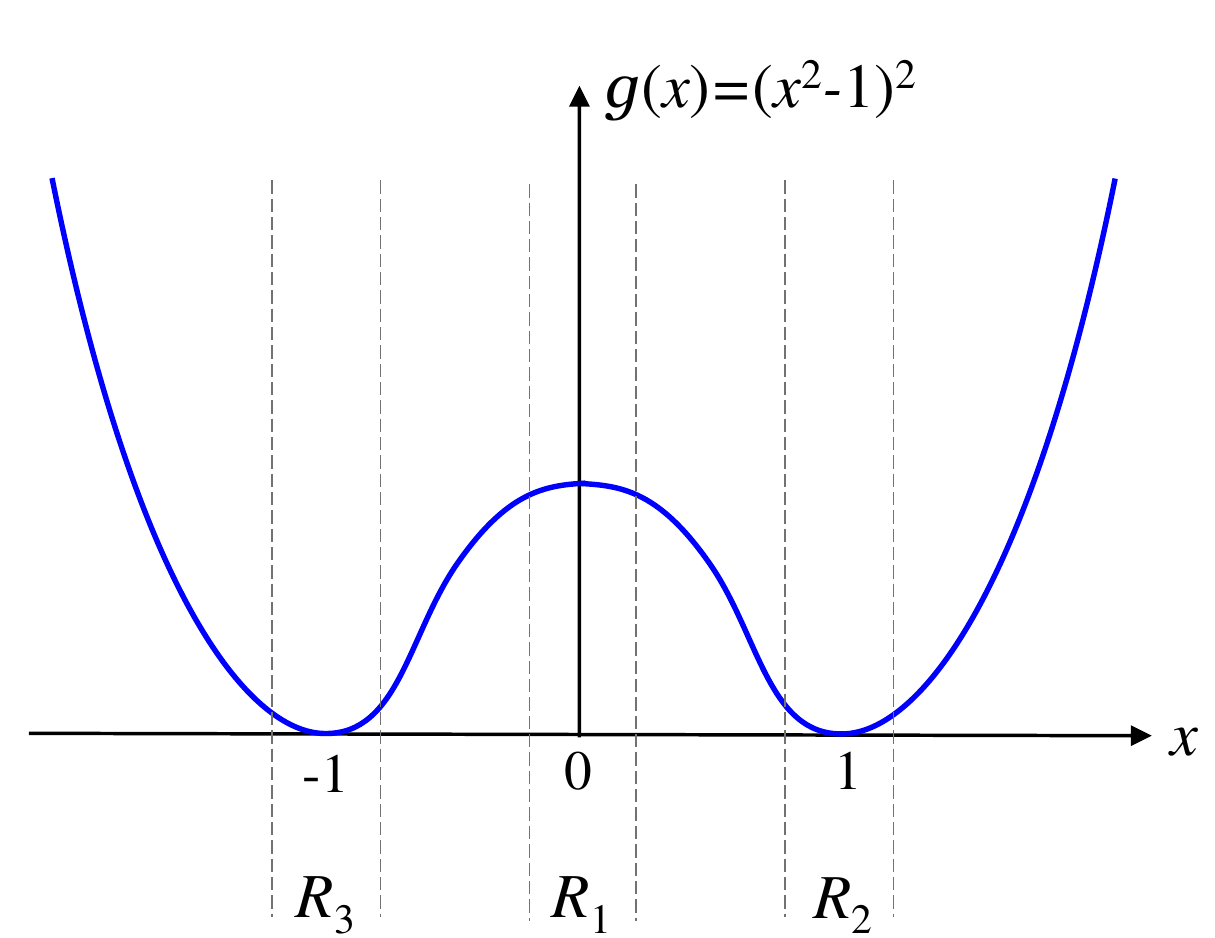}
	\end{center}
	\caption{Example of a one-dimensional strict saddle function: $ g(x) = (x^2 - 1)^2. $ This function has two local minima $ x^{\star} = 1 $ and $ x^{\star} =-1 $ that are also global minima, as well as a local maximum $ \bar{x}=0 $. In the region $ R_1 $, the function satisfies  $ \lambda_{\min} \left( \nabla^2 g(x)\right) \le -\gamma <0 $. In the regions $ R_2 $ and $ R_3 $, the function satisfies $ \twonorm{x - x^{\star}} \le \zeta$ for one of the local minima $ x^{\star} $. On the rest of the real line, the function satisfies $ | \nabla g(x) | \ge \epsilon >0 $. }\label{fig:strict_saddle}
\end{figure}

\begin{figure}[t]
	\begin{center}
		\hspace{-0.1in}\includegraphics[width=0.49\textwidth]{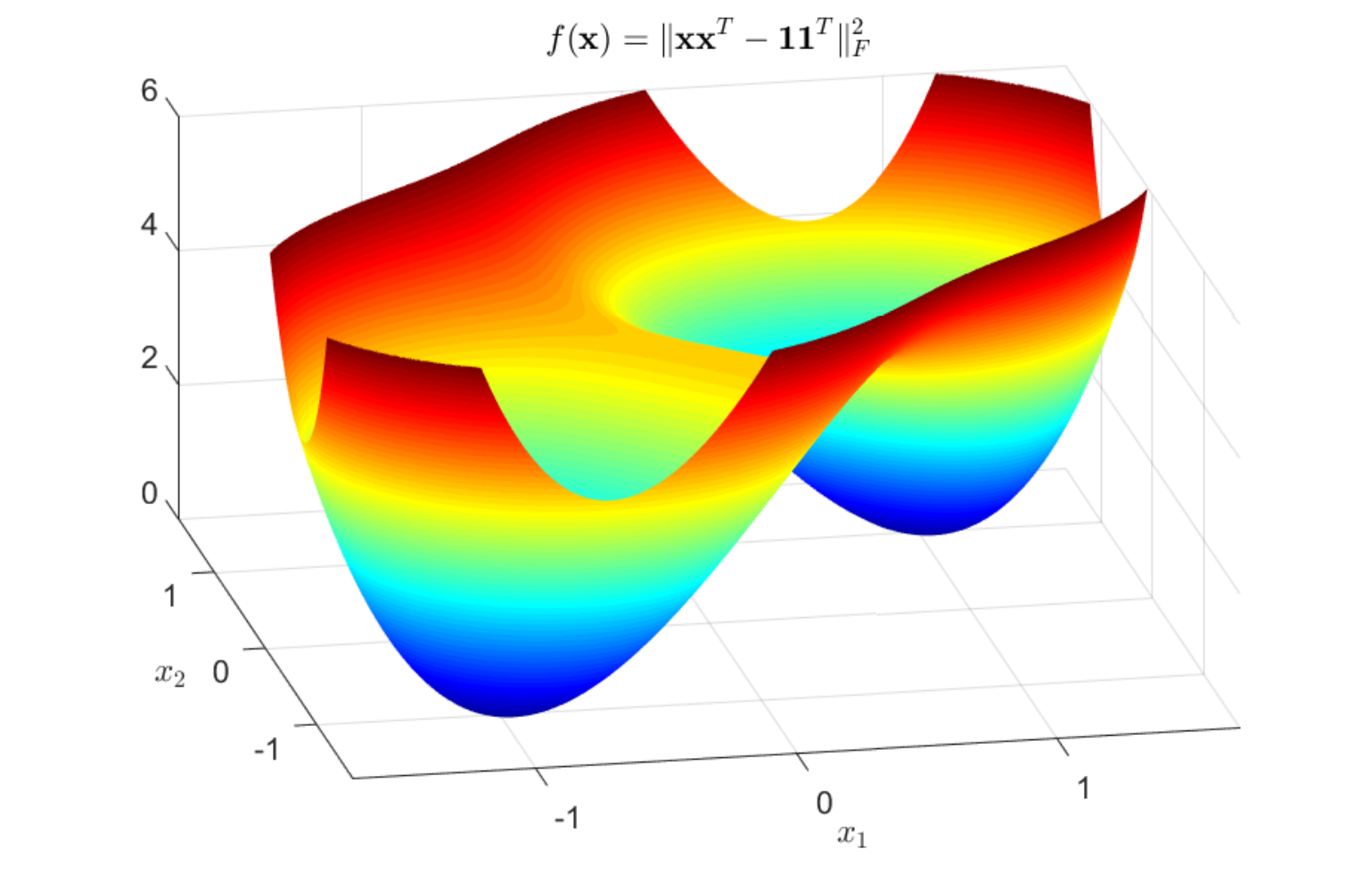}
	\end{center}
	\caption{Example of a two-dimensional strict saddle function $g(x) = \| \bx \bx^T - \mathbf{1}\mathbf{1}^T \|_\mathrm{F}^2 $, where $ \bx=[x_1, x_2]^T $ and $ \mathbf{1}=[1,1]^T $. This function has two local minima $ \bx^{\star} = [1,1]^T $ and $ \bx^{\star}=[-1,-1]^T $, and a strict saddle point $ \bx_{\text{saddle}} = [0,0]^T $. The Hessian $ \nabla^2 g(\bx_{\text{saddle}}) $ at the saddle point has a strictly negative eigenvalue with the eigenvector $ \left[ \frac{1}{\sqrt{2}}, \frac{1}{\sqrt{2}} \right]^T $, which corresponds to the descent direction $\bx^{\star} - \bx_{\text{saddle}} $. }\label{fig:strict_saddle2}
\end{figure}

In Figure~\ref{fig:strict_saddle} and Figure~\ref{fig:strict_saddle2} we provide examples of a one-dimensional function and a two-dimensional function, respectively, that satisfy the strict saddle property in Definition~\ref{def:strict_saddle}.
Intuitively, the strict saddle property of $ g(\bx) $ ensures that whenever one is not already close to a local minimum, the current solution will have either a large gradient, or a descent direction due to the Hessian having a negative eigenvalue. Therefore, any local search algorithms that are capable of finding such a descent direction, will make progress in decreasing the value of $g (\bx)$ and eventually converge to a local minimum. Many algorithms have been shown to enjoy this property, include cubic regularization~\cite{nesterov2006cubic}, trust-region algorithms~\cite{sun2015complete}, stochastic gradient descent~\cite{ge2015escaping}, and other more recent variants~\cite{jin2017escape,carmon2016accelerated,agarwal2016linear}.
In Algorithm~\ref{alg:pgd}, we describe one such algorithm, namely the Perturbed Gradient Descent (PGD) algorithm from~\cite{jin2017escape}. PGD is based on the standard gradient descent algorithm with the following additional steps: (i) when the gradient is small, indicating potential closeness to a saddle point, PGD adds a random perturbation to the current iterate (which is done at most once every $ t_{\text{thres}} $ iterations); (ii) if the last perturbation occurs $ t_{\text{thres}} $ iterations ago and the function value does not decrease sufficiently since, then PGD terminates and outputs the iterate before the last perturbation.
\begin{algorithm}[t]
	\caption{Perturbed Gradient Descent~\cite{jin2017escape}}
	\label{alg:pgd}
	\begin{algorithmic}
		
		\STATE \textbf{{Input}}: algorithm parameters $ d_{\text{thres}},  t_{\text{thres}}, g_{\text{thers}}, \stepit{}, R $;
		
		\STATE \textbf{{Initialization}}: Let $\bx_0=\boldsymbol{0}$;
		
		\STATE \textbf{for} $t=0,1,2,\ldots $ \textbf{do}
		
		\STATE 
		\begin{enumerate}
			\item \textbf{if} $ \| \nabla g(\bx_t) \|_2 \le d_{\text{thres}} $ and $ t-t_{\text{last}} > t_{\text{thres}} $ \textbf{then}
			
			~~~~$ \bx_{\text{last}} = \bx_{t} $, $ t_{\text{last}} = t $;
			
			~~~~$ \bx_{t} = \bx_{\text{last}} + \mathbf{\xi}$, where $ \mathbf{\xi} $ is sampled uniformly from the unit ball centered at zero with radius $ R $;
			
			\item \textbf{if} $ t-t_{\text{last}} = t_{\text{thres}} $ and $ g(\bx_t) -g(\bx_{\text{last}}) > - g_{\text{thres}} $ \textbf{then}
			
			~~~~\textbf{return} $ \bx_{\text{last}} $;
			
			\item $ \bx_{t+1} = \bx_{t} - \eta \nabla g(\bx_{t}) $.
		\end{enumerate}
		
	\end{algorithmic} 
\end{algorithm}

We do not delve further into the details of these saddle-escaping algorithms, as their parameter choices and run-time guarantees are somewhat technical. Rather, for the purpose of analyzing low-rank matrix estimation problems, we simply rely on the existence of such algorithms and the fact that their running time depends polynomially on the problem parameters. This is summarized in the following theorem, which abstracts out the key results in the work cited above.
\begin{theorem}[Optimizing strict saddle functions]
	\label{thm:saddle_convergence}
	Assume that $ g: \real^N \to \real $ is $ \beta $-smooth and $ (\epsilon, \gamma, \zeta) $-strict saddle.
	There exist  algorithms (such as PGD in Algorithm~\ref{alg:pgd} with appropriate choices of the parameters) that output a solution that is $ \zeta $-close to a local minimum of $ g $, with the required number of iterations upper bounded by a polynomial function of $ N, \beta, 1/\epsilon, \gamma $ and $ 1/\zeta $.
\end{theorem}

Specializing to the low-rank matrix estimation problem, it remains to verify that (i) the loss function defined on $ (\ltheta, \rtheta) $ is strict saddle and (ii) all its local minima are ``good'', in the sense that they correspond to a low-rank matrix equal to (in the noisy case, close to) the true matrix $ \bX $. To this end, we consider the same loss function as before for matrix sensing:
\[ g_{\mathcal{A}}(\bL,\bR)  = f_{\mathcal{A}}(\bL,\bR),
\]
where $ f_{\mathcal{A}}$ is defined in \eqref{eq:loss_sensing}. For matrix completion, we consider a regularized  loss function:
\begin{align*}
g_{\Omega}(\bU, \bV) = f_{\Omega}(\bL,\bR) + \lambda Q_{\alpha}(\bL, \bR), 
\end{align*}
where $f_\Omega$ is given in \eqref{eq:obj_proj_grad}, the regularizer is given by \cite{sun2014mc}
\begin{align*}
Q_{\alpha}(\bL, \bR) =   \sum_{i=1}^{n_1} \big( \twonorm{\be_i^T \bL } -\alpha \big)_+^4 +   \sum_{j=1}^{n_2} \big( \twonorm{\be_j^T \bR } -\alpha \big)_+^4,
\end{align*}
$\alpha$ and $\lambda$ are regularization parameters, and $(x)_{+}=\max\{x,0\}$. The regularization plays a similar role as the projections $ \Proj_{\LCons }, \Proj_{\RCons} $ previously used: it encourages incoherence of $ \bL$ and $ \bR $. Replacing projections with regularization leads to an unconstrained formulation that fits into the strict-saddle framework above. The follow theorems show that these loss functions indeed have the desired strict-saddle property with high probability under sample complexity conditions similar to before.
\begin{theorem}[Matrix Sensing] \cite{bhojanapalli2016global, ge2017no}
	\label{thm:saddle_sensing}
	Suppose that the measurement operator 	$\Xmap$ satisfy RIP-$\ell_2/\ell_2$ with parameter $ \ripparam{2\rdim} =\max\{\text{\underbar{$\delta$}}_{2r}, \bar{\delta}_{r}\}  = \frac{1}{20} $. For any $ \epsilon>0 $, the above loss function~$ g_{\mathcal{A}} $ satisfies the following: (i) it is $ \big( \epsilon, \Omega(\sigma_\rdim), O(\frac{\epsilon}{\sigma_\rdim}) \big) $-strict saddle, and (ii) all its local minima satisfying $ \CROSS{\ltheta}{\rtheta}= \bX$.
\end{theorem}

\begin{theorem}[Matrix Completion] \cite{GeLeeMa2016,ge2017no}
	\label{thm:saddle_mc}
	Suppose that the observation probability satisfies $ \pobs \ge \Omega \big( \frac{\inco^4 \rdim^6  \log (n_1+n_2)}{(n_1+n_2)} \big) $, and one chooses $ \alpha^2 = \Theta\big( \frac{\inco \rdim \sigma_1}{(n_1+n_2)} \big) $ and $ \lambda = \Theta \big( \frac{(n_1+n_2)}{\inco \rdim} \big) $. Then, with probability at least $ 1-(n_1+n_2)^{-1}$, the above  loss function~$ g_{\Omega} $ satisfies the following: (i) it is $ \big( \epsilon, \Omega(\sigma_\rdim), \order(\frac{\epsilon}{\sigma_\rdim}) \big) $-strict saddle for any $ \epsilon \le \text{poly} (\frac{\inco^4 \rdim^4 \sigma^{4}_1}{(n_1+n_2)^2} ) $, and (ii) all its local minima satisfy $ \CROSS{\ltheta}{\rtheta}  = \bX$.
\end{theorem}

Combining Theorem~\ref{thm:saddle_convergence} with Theorem~\ref{thm:saddle_sensing} and Theorem~\ref{thm:saddle_mc}, we conclude that iterative algorithms optimizing over the factor variables $ (\ltheta, \rtheta) $ converge globally to some pair satisfying $ \ltheta \rtheta^T = \bX $ in a polynomially number of iterations from any arbitrary initial solutions, as long as they can escape saddle points. We refer the readers to~\cite{ge2017no,jin2017escape} for more discussions.
 
\subsection{Perspectives} 

Combining the discussions in the last two subsections, we obtain the following general picture for low-rank matrix estimation reformulated in the factor space of $ (\ltheta, \rtheta) $: 
\begin{itemize}
	\item All the local minima of the loss function are in fact global minima, and correspond to some factorization  $  \CROSS{\ltheta }{\rtheta } $ of the true low-rank matrix $ \bX  $.
	\item In a neighborhood of each global minimum $ (\bL, \bR) $, the loss function is essentially strongly convex, and has no saddle point. Within this neighborhood, gradient descent and other iterative algorithms converge geometrically.
	\item Any point outside such neighborhoods either has a strictly positive gradient, or is a saddle point with a descent direction corresponding to a strictly negative eigenvalue of the Hessian.
	\item Iterative algorithms escape all saddle points and enter a neighborhood of the global minima in polynomial time. Alternatively, one can find a solution in this neighborhood by performing one SVD of a matrix appropriately constructed from the observations.
\end{itemize}

Comparing the two approaches to non-convex matrix estimation discussed in the last two subsections, we also see that each of them has its own strengths. The first approach, taken in Section~\ref{sec:local}, focuses on the convergence of algorithms with a proper initialization procedure. This approach immediately leads to simple, efficient algorithms, with provably geometric convergence and linear time-complexity. It also readily extends to problems that have additional structural constraints (such as sparsity, Hankel and discrete structures),  or involve more complicated loss functions (such as robust PCA and matrix completion with quantized observations), some of which may involve a non-smooth loss function whose Hessian is not defined. However, finding a good initialization scheme is non-trivial, and in some settings is actually the harder part of the problem. The second approach, taken in Section~\ref{sec:global}, instead focuses on the global geometric landscape of the problem. This approach is conceptually elegant, nicely decomposing the geometric aspect (properties of the local minima) and the algorithmic aspect (how to find the local minima) of the problem. Computationally, it eliminates the need of careful initialization, but the resulting run-time guarantees are somewhat weaker, which may be super-linear in the dimensions. Of course, we made the distinction between these two approaches mostly for ease of review of state-of-the-art;  given the rapid developments in this area, we expect that both approaches will be improved, expanded, and eventually merged.

Before concluding this section, we add that there is a deeper reason for low-rank matrix estimation being such a benign nonconvex problem. The loss function of the low-rank matrix estimation problem can often be viewed, in a certain precise sense,  as a perturbed version of the objective function of PCA, i.e., finding the best rank-$r$ approximation in Frobenius norm in a factorized form:
\begin{align}\label{eq:best_rank_r}
\min_{\bL \in \in \real^{n_1 \times \rdim},\bR \in \real^{n_2 \times \rdim}} \| \CROSS{\ltheta}{\rtheta} - \bX\|_{\mathrm{F}}^2.
\end{align} 
For example, the matrix completion loss function~\eqref{eq:obj_proj_grad}, with the regularization omitted, is exactly equal to the above objective \emph{in expectation}. The PCA problem~\eqref{eq:best_rank_r} is arguably the most well-understood tractable non-convex problem: in addition to having a closed-form solution \eqref{close_form}, this problem satisfies all the geometric properties mentioned in the last two subsections~\cite{yang1995projection,li2016symmetry}; in particular, its local minima and saddle points can be expressed in terms of the top and non-top eigen components of $ \bX $, respectively. Under the probabilistic or RIP assumptions on the sensing operators, the geometric and algorithmic properties of the PCA problem~\eqref{eq:best_rank_r} are essentially preserved under incomplete observations, with high probability, as long as the issue of incoherence is appropriately accounted for.

\section{Structured Low-Rank Matrix Estimation}\label{sec_structured}

In many applications, the low-rank matrix estimation problems possess additional structures that need to be carefully exploited, and we present two such examples in this section: Hankel matrix completion, and the  recovery of clustering matrices. 

\begin{figure*}[t]
	\begin{center}
		\begin{tabular}{cccc}
	 \hspace{-0.15in}	\includegraphics[width=0.25\textwidth ]{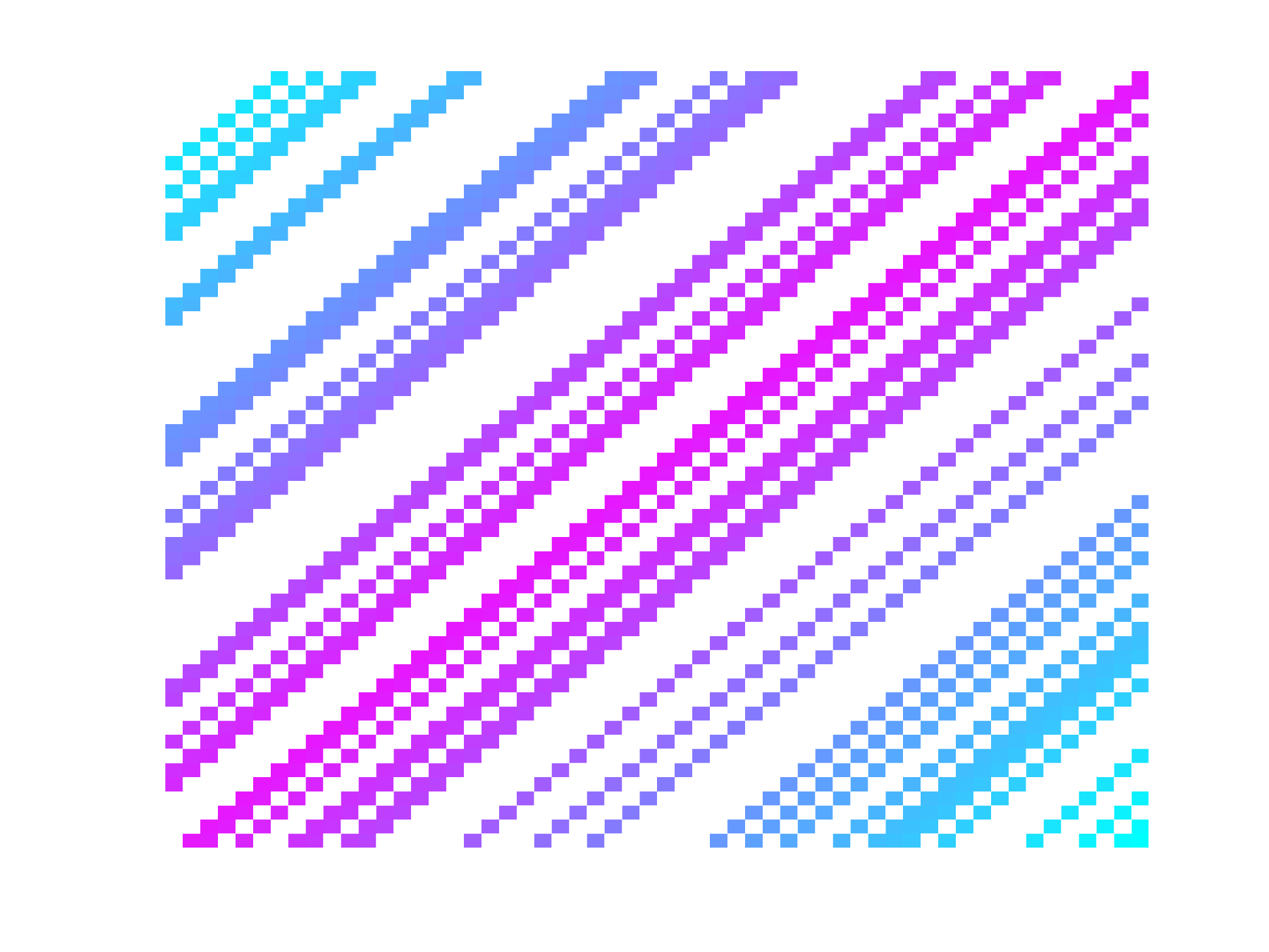} & \hspace{-0.15in}\includegraphics[width=0.25\textwidth ]{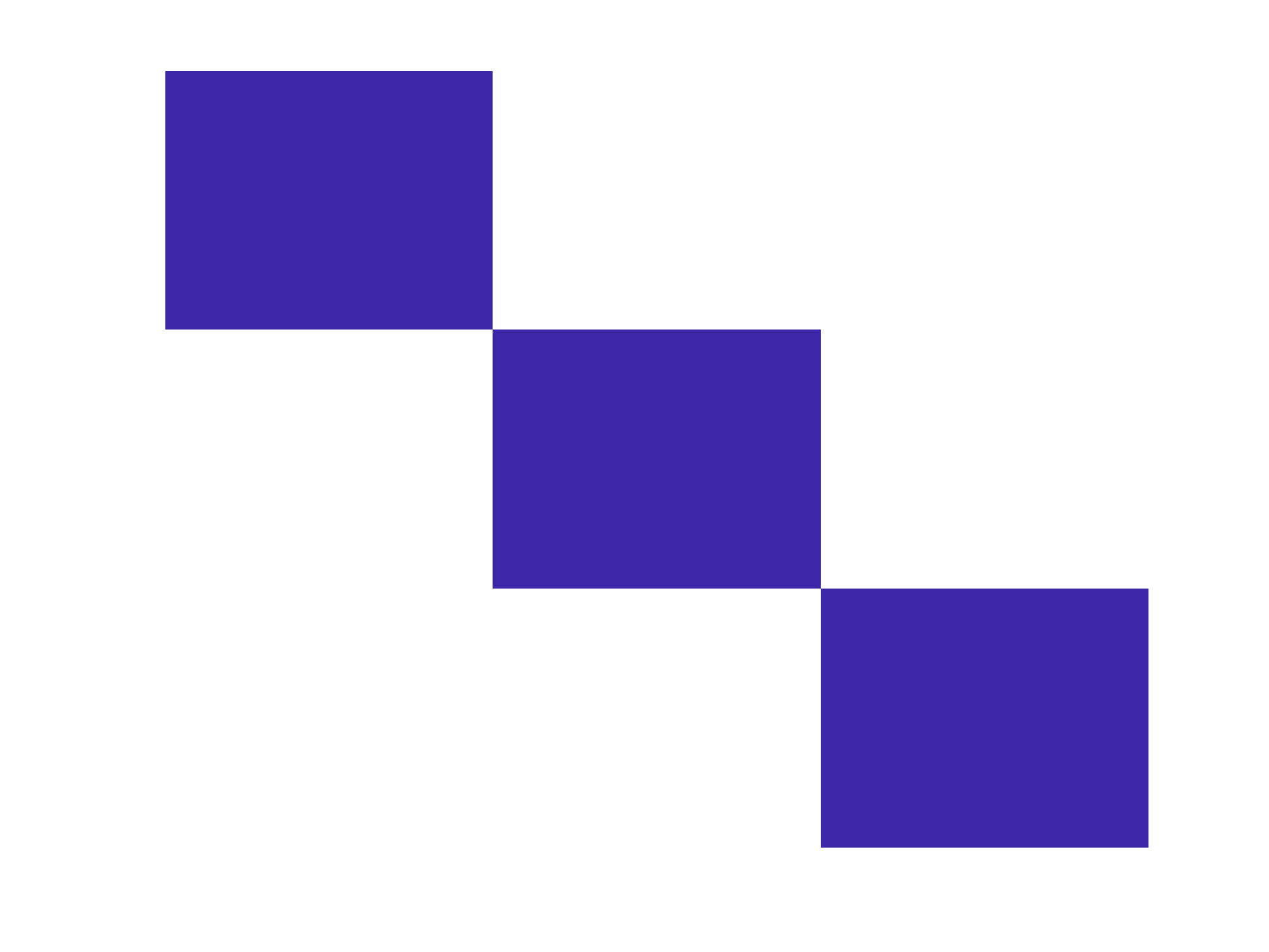}  &\hspace{-0.15in}\includegraphics[width=0.25\textwidth ]{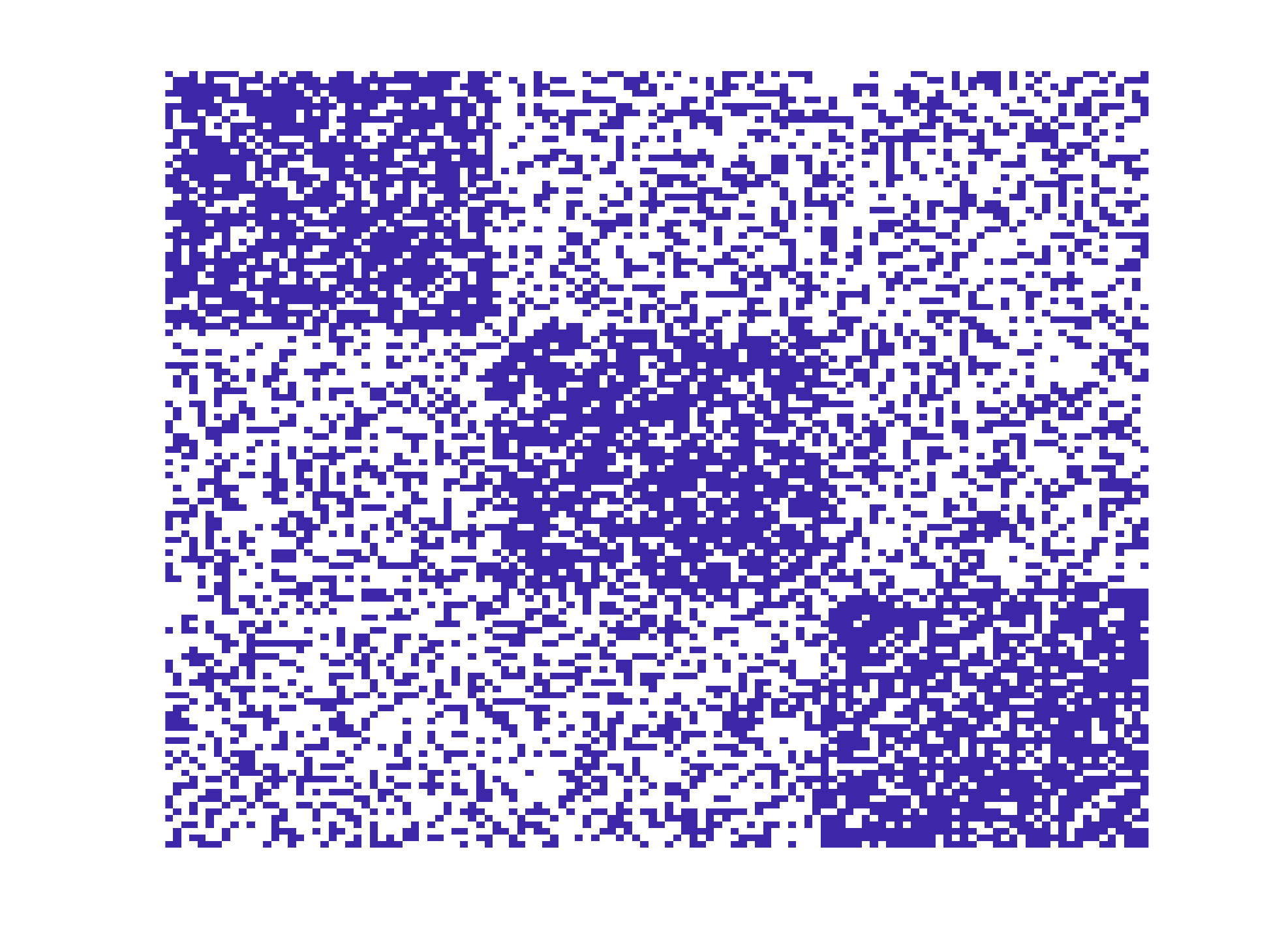} & \hspace{-0.15in}\includegraphics[width=0.25\textwidth ]{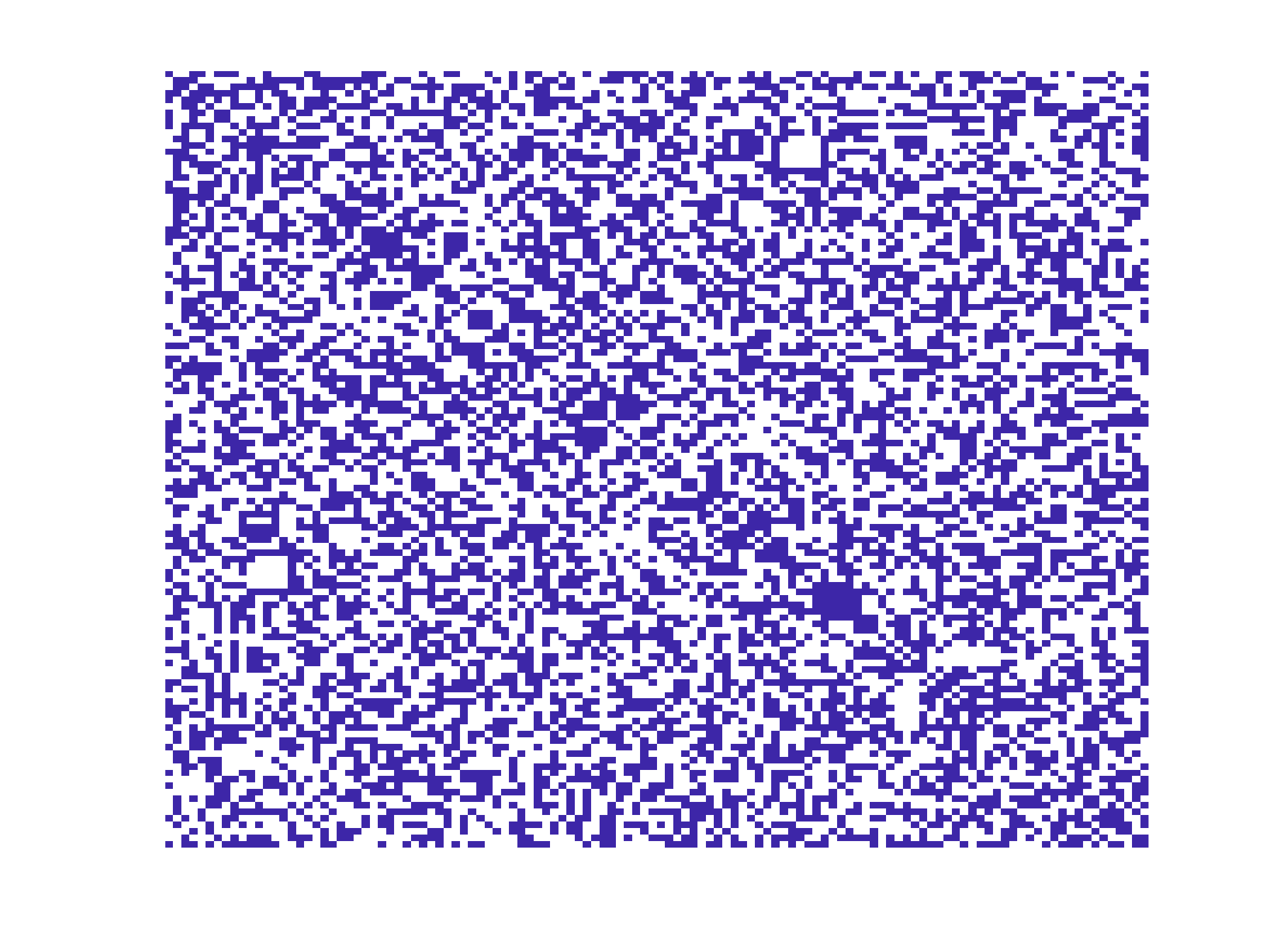} \vspace{-0.1in}\\ 
		 \hspace{-0.15in}(a) & \hspace{-0.15in} (b) & \hspace{-0.15in} (c) &  \hspace{-0.15in}(d)
		\end{tabular}
	\end{center}
	\caption{Illustration of structured matrices considered in Section~\ref{sec_structured}. (a) The observation pattern in a Hankel matrix completion problem. (b) The cluster matrix, and (c) the affinity matrix in a cluster matrix recovery problem, when the nodes are ordered according to the cluster structure. (d) Same as (c), except that the nodes are randomly permuted. }\label{fig:cluster_matrix}
\end{figure*}

\subsection{Hankel Matrix Completion} \label{sec:hankel}

Imagine that one is interested in estimating the spectrum of time series, or the direction-of-arrivals from returns from a sensor array. One can model the signal of interest as a weighted sum of complex exponentials, i.e.,
\begin{equation}\label{harmonic}
\bx = \sum_{i=1}^r  c_i \boldsymbol{v}(z_i), 
\end{equation}
where $c_i\in\mathbb{C}$ represents the complex amplitude, $z_i \in \mathbb{C}$ is the complex frequency, and $ \boldsymbol{v}(z_i) $ is of the form
\begin{equation}
\boldsymbol{v}(z) = \begin{bmatrix}
1 & z & \cdots & z^{n-1}
\end{bmatrix}^T \in\mathbb{C}^n.
\end{equation} 
One can view the atom $\boldsymbol{v}(z)$ as an eigenvector of a linear time-invariant system, with $z$ being the corresponding pole. 

Spectral compressed sensing concerns with the problem of recovering the signal $\bx$ from only a subset of its entries. This is an important problem in super resolution~\cite{chen2014robust}, MRI imaging~\cite{jin2016general}, and system identification~\cite{Fazel2003Hankel}. Denoting the index set of the observed entries by $\Omega\subset\{1,\ldots,n\}$, our goal is to recover $\bx$ given $\mathcal{P}_\Omega(\bx)$. In the literature of compressed sensing, this problem is typically approached by first discretizing the parameter space of $z$, and then solving an $\ell_1$ minimization problem as in standard sparse recovery. However, this approach is sensitive to the discretization used, and cannot resolve parameters that fall off the grid~\cite{chi2011sensitivity}.

It turns out that under certain mild conditions, this problem can be solved \emph{exactly} as a structured matrix completion problem, without assuming any discretization/grid. The insight lies in exploiting the shift invariance property embedded in the structure of complex harmonics. This is done by constructing an $n_1$-by-$ (n-n_1+1)$ Hankel matrix spanned by the signal vector $\bx\in\mathbb{C}^{n}$ as
\begin{equation}
\mathcal{H}(\bx) =\begin{bmatrix}
x_1 &  x_2  & &\\
x_2 &  & \iddots& \\
\vdots & \iddots& & \\
x_{n_1} & x_{n_1+1}& \cdots & x_{n}
\end{bmatrix},
\end{equation}
where $n_1$ is commonly selected as $\lfloor n/2 \rfloor$ to make the matrix $\mathcal{H}(\bx) $ as square as possible.
The important observation is that $\mathcal{H}(\bx)$ admits the following low-rank decomposition:
\begin{equation}
\mathcal{H}(\bx) =\boldsymbol{V}_{n_1}\boldsymbol{C}\boldsymbol{V}_{n-n_1+1}^{T},
\end{equation}
where
\begin{align}\label{eq:Vn1}
\boldsymbol{V}_{n_1} =  \begin{bmatrix}
1 & 1 & \cdots  & 1 \\
z_1 & z_2 & \cdots & z_r \\
\vdots & \vdots & \vdots & \vdots\\
z_1^{n_1-1} & z_2^{n_1-1} & \cdots & z_r^{n_1-1} 
\end{bmatrix},
\end{align}
$\boldsymbol{C}=\diag[c_1,c_2,\ldots,c_r]$, and $\boldsymbol{V}_{n-n_1+1}$ is defined in a way similar to \eqref{eq:Vn1}.
This decomposition shows that $\mbox{rank}(\mathcal{H}(\boldsymbol{x}))\leq r$, and equality holds when all the poles are distinct. This representation of $ \bx $ as a structured low-rank matrix can be leveraged to facilitate recovery of the un-measured entries of~$ \bx $. In particular, one can try to recover the missing measurements by seeking a Hankel matrix with the smallest nuclear norm and consistent with the available measurements. This idea gives rise to the following algorithm, termed Enhanced Matrix Completion (EMaC) \cite{chen2014robust}:
\begin{equation}
\min_{\bg\in\mathbb{C}^n} \|\mathcal{H}(\bg) \|_* \quad \mbox{subject to} \quad \mathcal{P}_{\Omega}(\bg)=\mathcal{P}_{\Omega}(\bx).
\end{equation}
Figure~\ref{fig:cluster_matrix} (a) illustrates the observation pattern in a Hankel matrix recovery problem, which is highly structured. 
 
Under the parametric model \eqref{harmonic}, we define a new notion of incoherence that bears an interesting physical interpretation. Let the Dirichlet kernel be 
\begin{equation}
\mathcal{D}_{n_1}( z):=\frac{1}{n_{1} }\left(\frac{1-z^{n_1}}{1-z}\right), \label{dirichlet}
\end{equation}
whose absolute value decays inverse proportionally with respect to $|z|$. Given $r$ poles, one can construct two $r\times r$ Gram matrices $ \boldsymbol{G}_{\text{L}} $ and $ \boldsymbol{G}_{\text{R}} $, corresponding to the column space and row space of $\mathcal{H}(\bx)$, where the entries of these matrices are specified by 
\begin{align*}
(\boldsymbol{G}_{\text{L}})_{i,l} & =\mathcal{D}_{n_1}( z_i -z_l ), \quad 1\leq i,l\leq r; \\
(\boldsymbol{G}_{\text{R}})_{i,l} & =\mathcal{D}_{n-n_1+1}( z_i -z_l), \quad 1\leq i,l\leq r.
\end{align*}
The incoherence parameter is then defined as follows.
\begin{definition}[\textbf{Incoherence}]
	The incoherence parameter of a signal $\bx$ of the form~\eqref{harmonic} is defined as the smallest number~$\mu$ satisfying the bounds
\begin{equation}
\sigma_{\min}\left(\boldsymbol{G}_{\mathrm{L}}\right)\geq\frac{1}{\mu}
\quad\text{and}\quad
\sigma_{\min}\left(\boldsymbol{G}_{\mathrm{R}}\right)\geq\frac{1}{\mu},
\label{eq:LeastSV_G}
\end{equation}
where $\sigma_{\min}\left(\boldsymbol{G}_{\mathrm{L}}\right)$ and
$\sigma_{\min}\left(\boldsymbol{G}_{\mathrm{R}}\right)$ denote
the smallest singular values of $\boldsymbol{G}_{\mathrm{L}}$ and $\boldsymbol{G}_{\mathrm{R}}$,
respectively. 
\end{definition}

If all poles are well-separated by $2/n$, the incoherence parameter $ \mu $ can be bounded by a small constant \cite{liao2016music}. As the poles get closer, the Gram matrices become poorly-conditioned, resulting in a large $ \mu $. Therefore, the incoherence parameter provides a measure of the hardness of the recovery problem in terms of the relative positions of the poles. The theorem below summarizes the performance guarantees of the EMaC algorithm.
\begin{theorem}[\cite{chen2014robust}]
	\label{thm:hankel}
	Suppose that each entry of $ \bx $ is observed independently with probability $p$. As long as 
	\[ p\geq C\frac{\mu r\log^4 n}{n},
	\] 
	for some sufficiently large constant $C$, the signal $\bx$ can be exactly recovered with high probability via EMaC. 
\end{theorem}

Theorem~\ref{thm:hankel} suggests that a Hankel-structured low-rank matrix can be faithfully recovered using a number of measurements much smaller than its dimension $n$. Recently it has been shown that Hankel matrix completion can also be efficiently solved using the non-convex \BM factorization and projected gradient descent approach described in Section~\ref{sec_nonconvex}, under similar conditions~\cite{cai2017spectral}. Similar results can be obtained for block Hankel or Toeplitz low-rank matrix completion (for multi-dimensional data) as well. Interestingly, if the Toeplitz matrix is additionally positive-semidefinite, the incoherence condition can be relaxed by exploring the connection to Carath\'eodory's theorem; see \cite{chen2015exact,qiao2017gridless}.

\subsection{Cluster Matrices}
\label{sec:clustering} 

Suppose that we are given an affinity matrix $ \Yout \in \real^{\usedim \times \usedim}$ between $ \usedim $ nodes, where $ Y_{ij} $ is a measure of the pairwise similarity/affinity between nodes $ i $ and $ j $. For example, each $ Y_{ij} $ may be an indicator of the friendship of two Facebook users, or the similarity between two movies on Netflix. Our goal is to partition these $ \usedim $ nodes into several clusters such that nodes within the same clusters have high affinity values. This problem is known as \emph{clustering} or \emph{community detection}. 

One may represent an ideal partition by a so-called \emph{cluster matrix} $ \boldsymbol{X}^{\star} \in \{0,1\}^{\usedim \times \usedim} $ defined as
\begin{align*}
X^{\star}_{ij} = 
	\begin{cases}
	1 & \text{if nodes $ i $ and $ j $ are in the same cluster}, \\
	0 & \text{if nodes $ i $ and $ j $ are in different clusters}.
	\end{cases}
\end{align*}
With an appropriate ordering of the rows and columns, the matrix $ \boldsymbol{X}^{\star} $ takes the form of a block-diagonal matrix:
\begin{align*}
\boldsymbol{X}^{\star} = 
	\begin{bmatrix}
	\OneMat_{\csize_1 \times \csize_1} & & &\\
	& \OneMat_{\csize_2 \times \csize_2} & &\\
	& & \ddots & \\
	& & & \OneMat_{\csize_\rdim \times \csize_\rdim}
	\end{bmatrix},
\end{align*}
where $ \rdim $ is the number of clusters, $ \csize_k $ is the size of the $ k $-th cluster, and $ \OneMat_{\csize \times \csize} $ denotes the $ \csize $-by-$ \csize $ all-one matrix.
It is clear that the rank of $ \boldsymbol{X}^{\star} $ is equal to the number of clusters~$ \rdim $. Moreover, the matrix $ \boldsymbol{X}^{\star} $ has several additional structural properties: it is binary, block-diagonal, positive-semidefinite, and has all diagonal entries equal to one.

The fact that the nodes in the same cluster tend to have high affinity values, can be captured by the model
$
Y_{ij} = X^{\star}_{ij} + W_{ij},
$
where $ W_{ij} $ is some form of noise that encapsulates the inherent randomness/uncertainty in the pairwise affinity measure. In many applications, the affinity values between some pairs of nodes are unknown, or are costly to measure, in which case we only observe a subset $ \Obs $ of the entries of $ \Yout $. Under this setup, the clustering problem can be cast as a noisy low-rank matrix completion problem, and the algorithms and theory in the last two sections can be immediately applied. 

Notably, one can take advantage of the additional structures of the cluster matrix $ \boldsymbol{X}^{\star} $ to obtain stronger performance guarantees. In particular, it is sometimes possible to \emph{recover $ \boldsymbol{X}^{\star} $ exactly even in the presence of noise}. 
We briefly review one such result from \cite{chen2012sparseclustering}. 
Consider the setting where the noise term $ W_{ij} $ is such that $ Y_{ij} \sim \text{Bernoulli}(\pin) $ if $ X^{\star}_{ij} = 1 $ and  $ Y_{ij} \sim \text{Bernoulli}(\qout) $ if $ X^{\star}_{ij} = 0$, where $ \pin > \qout $; in this case, nodes in the same clusters have a higher probability of having a high (non-zero) affinity value. Figure~\ref{fig:cluster_matrix} (b) and (c) illustrate the cluster matrix and the affinity matrix where the nodes are ordered according to the cluster structure, and (d) shows the affinity matrix except that the nodes are randomly permuted, as how it is typically observed in practice. As before, we assume that each entry of $ \Yout $ is observed with some probability $ \pobs $. This model is sometimes referred to as the \emph{(censored) stochastic block model} or \emph{planted partition model} in the literature~\cite{holland83,condon2001algorithms}.
For this model, we consider the maximum likelihood estimator of $ \boldsymbol{X}^{\star} $, and derive a convex relaxation of it by replacing the non-convex constraints on $ \boldsymbol{X}^{\star} $ (low-rank, binary and block-diagonal) with a nuclear norm regularizer and linear inequality constraints $ X^{\star}_{ij} \in [0,1] $. Doing so leads to the semidefinite program 
\begin{equation}
\label{eq:clusetring_convex}
\begin{aligned}
\min_{\bX \in \real^{n\times n}} \;\;  & -\alpha \langle \ProjObs(\bY), \bX \rangle + \alpha^{-1} \langle \OneMat - \ProjObs(\bY), \bX \rangle  + 48\sqrt{n} \| \bX \|_*\\
\text{subject to} \;\; & X_{ij} \in [0,1], \forall i,j,
\end{aligned}
\end{equation}
where $ \alpha = \sqrt{\frac{2-\pobs(\pin+\qout)}{\pobs(\pin +\qout)}}  $ and $ \OneMat = \OneMat_{n\times n} $ is the $ n\times n$ all-one matrix; see~\cite{chen2012sparseclustering} for the details.
This approach enjoys the following guarantees.
\begin{theorem}[Corollary 3 in \cite{chen2012sparseclustering}]
	Suppose that the minimum cluster size is $ \csize_{\min} $. As long as
	\begin{align*}
	 \frac{\pobs (\pin - \qout)^2}{\pin}  \ge C \frac{\usedim \log^2 \usedim}{\csize_{\min}^2}
	\end{align*}
	for some constant $ C $, the convex relaxation~\eqref{eq:clusetring_convex} recovers $ \boldsymbol{X}^{\star} $ exactly as the unique minimizer with high probability.
\end{theorem}
In words, provided that the observation probability and the difference between $ \pin $ and $ \qout $ are large enough,  the solution of the convex relaxation formulation is guaranteed to have the structures of a cluster matrix and equal $ \boldsymbol{X}^{\star} $ exactly.

Clustering is a classical problem that has been studied extensively, with a huge body of literature. The above perspective of casting it as a low-rank matrix recovery problem is relatively recent, and proves to be fruitful, leading to a number of new algorithms and theoretical results. A detailed account of these developments is outside the scope of this article, and we refer the readers to the recent surveys~\cite{abbe2016recent,chen2013planted} and the references therein.

\section{Numerical Examples on MovieLens Data}\label{sec_simulations}

In this section, we showcase some numerical results of applying the matrix completion approach to a real dataset, namely, the MovieLens 100K Dataset \cite{harper2016movielens}. The dataset consists of $100,000$ ratings, taking values of $1,2,3,4$ or $5$, from $943$ users on $1682$ movies. We work with a popular version of the data that partitions the entire dataset `u.data' into a training set `ua.base' and a test set `ua.test', where the test set contains exactly $10$ ratings per user. We further delete two movies with no ratings available. Let $\bM$ denote the original incomplete rating matrix generated with rows corresponding to users and columns corresponding to movies, and let $ \Phi $ be the index set of the available ratings  contained in `u.data'. With this notation, we have $ |\Phi| = 10^{5} $, and $\bM$ is a $943$-by-$1680$ matrix where for each $ (i,j) \in \Phi $, $M_{ij} \in \{1,2,3,4,5\}$.  Further, let $\Omega$ and $\Omega^{c}$ denote the disjoint index sets of ratings for training and testing, respectively, with $\Omega \cup \Omega^{c} = \Phi$. In the experiment, we use the training data set $\Omega$ as the input to a matrix completion algorithm. The completed matrix produced by the algorithm, denoted by $\hat{\bM} \in\mathbb{R}^{943\times 1680}$, is used as an estimate of the unobserved ratings and will be evaluated on the test set $ \Omega^c $. 

We demonstrate performance of three matrix completion algorithms: accelerated proximal gradient \cite{toh2010accelerated}, singular value projection \cite{jain2010svp}, and bi-factored gradient descent \cite{park2016finding}. For these algorithms we use existing implementations with publicly available codes, and mostly adopt their default settings with only a few adjustments (detailed below) tailored to this specific dataset. For the error metric, we use the normalized mean absolute errors (NMAE) over the training set and test set, defined respectively as
\begin{align*}
\mathrm{NMAE}_{\text{train}} 
& = \frac{1}{(5-1)\left\vert \Omega \right\vert} \sum_{(i,j)\in \Omega} \left\vert \hat{M}_{ij} - M_{ij}\right\vert,\\
\mathrm{NMAE}_{\text{test}} 
& = \frac{1}{(5-1)\left\vert \Omega^{c} \right\vert} \sum_{(i,j)\in \Omega^{c}} \left\vert \hat{M}_{ij} - M_{i,j}\right\vert.
\end{align*}

\begin{figure*}[htp]
\centering 
\begin{tabular}{cc}
\includegraphics[height=1.6in,width=0.4\textwidth]{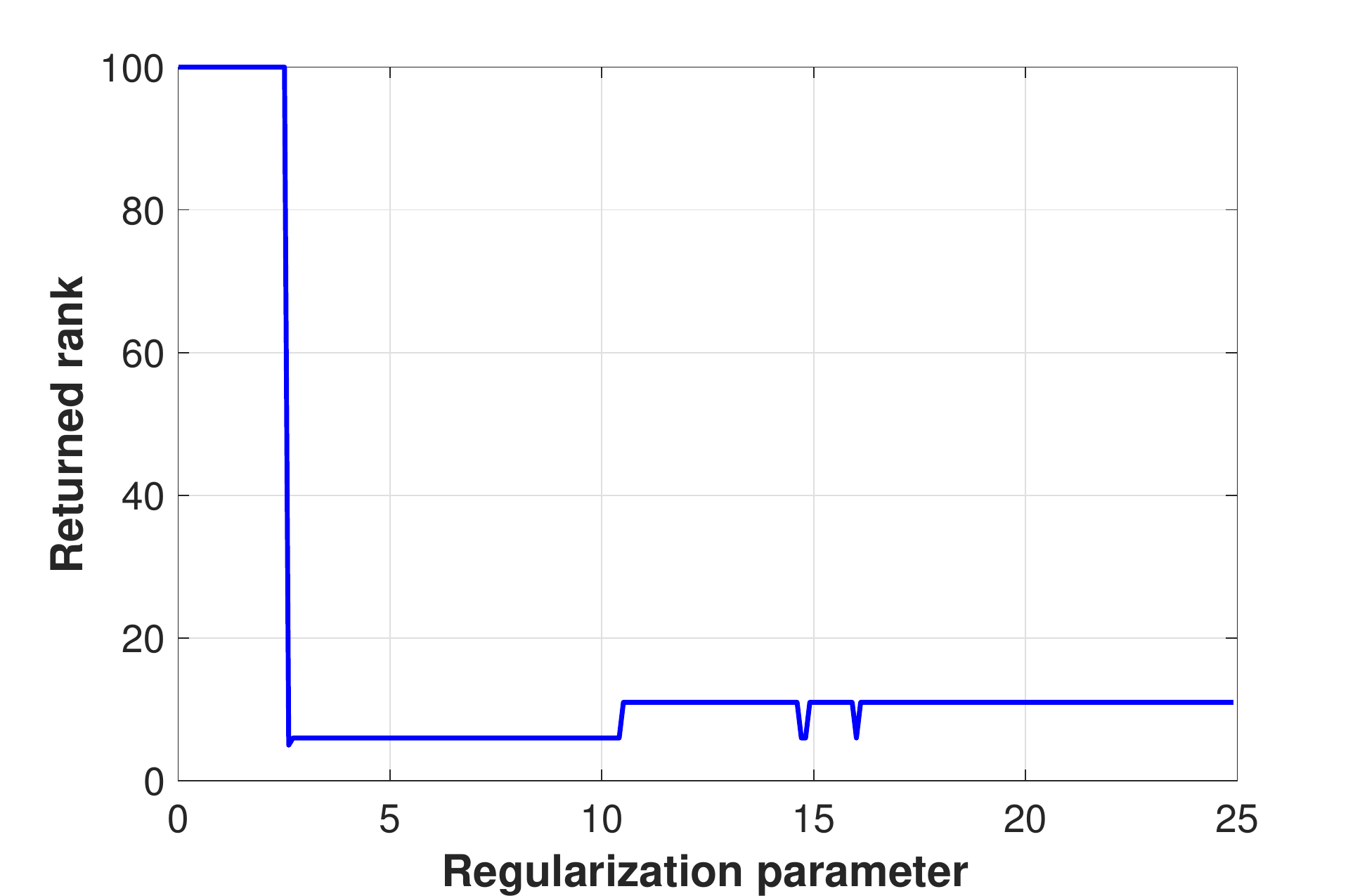} &\includegraphics[height=1.6in,width=0.4\textwidth]{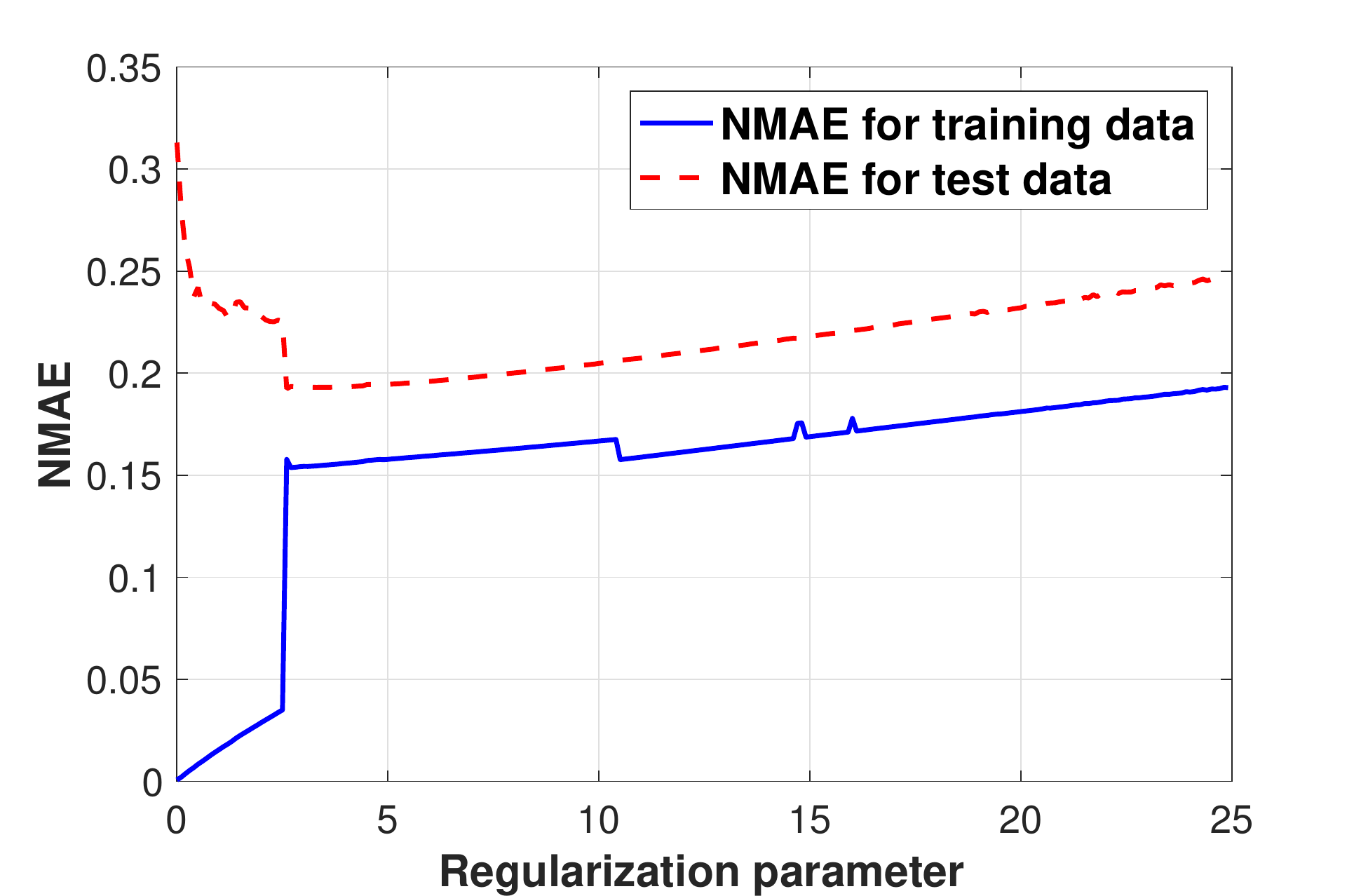} \\
(a) & (b)
\end{tabular}
\caption{Performance of APG on MovieLens 100K dataset: (a) The rank of estimated matrix via APG with respect to the regularization parameter. (b) The NMAEs for training data and test data via APG with respect to the regularization parameter.}\label{fig_main_MC_APG}
\end{figure*}

We first employ the accelerated proximal gradient (APG) algorithm proposed in \cite{toh2010accelerated}.\footnote{\url{http://www.math.nus.edu.sg/~mattohkc/NNLS.html}} We disable the adaptive updating of the regularization parameter in the implementation, and instead use a fixed one. We set the maximum rank to $100$ and the maximum number of iterations to $1500$. In our experiment, we observe that the algorithm usually meets the stop criteria after just a few of iterations, and hence stops early before reaching the maximum number of iterations. When using different values for the regularization parameter, APG outputs an estimate matrix $ \hat{\bM} $ with different ranks. Fig.~\ref{fig_main_MC_APG} (a) shows the relation between the regularization parameter and rank, and Fig.~\ref{fig_main_MC_APG} (b) shows NMAEs for the training data and test data against different ranks. The minimum NMAE on the test data is $0.1924$, which is achieved when the regularization parameter is set to $2.61$ with the rank of the estimate being $5$. 

\begin{figure}[htp]
\centering 
\includegraphics[height=1.6in,width=0.4\textwidth]{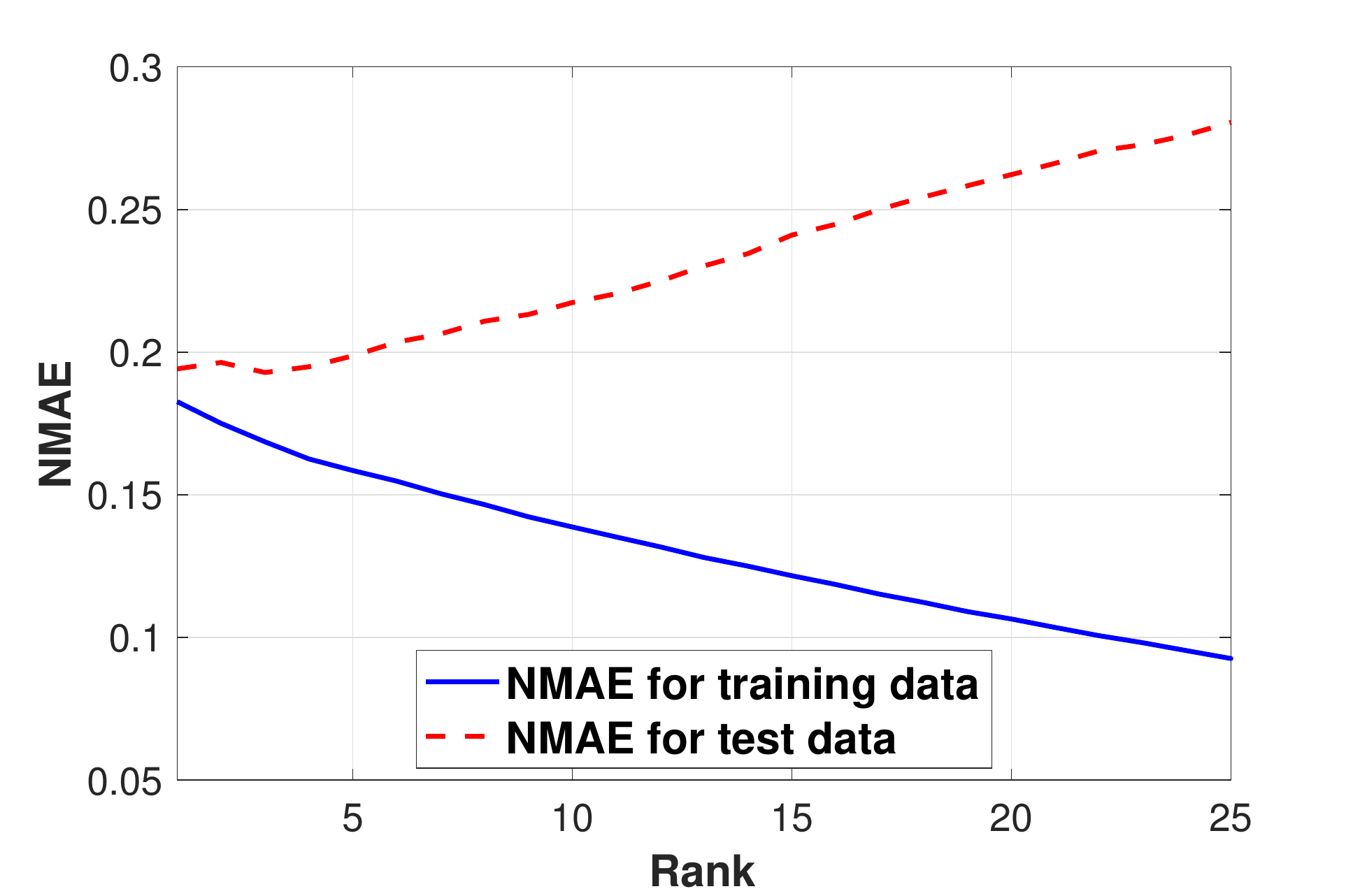}
\caption{Performance of SVP \cite{jain2010svp} on MovieLens 100K dataset: The NMAEs for training data and test data via SVP with respect to the rank.}
\label{fig_main_MC_SVP_NMAE_tog}
\end{figure}

We next consider the singular value projection (SVP) algorithm proposed in \cite{jain2010svp}.\footnote{\url{http://www.cs.utexas.edu/~pjain/svp/}} The stopping criteria are set as $\mathrm{tol} = 10^{-3}$ and $\mathrm{vtol} = 10^{-4}$, and the maximum number of iterations is set to $1000$. Again, the algorithm usually stops early before reaching the max iterations. The step size is chosen to be $\eta = 0.1 \times 3/(4\times p)$, where $p = |\Omega| /(943 \times 1680)$ is the fraction of available ratings in the training set. The rank of the estimate matrix is itself a user-specified tuning parameter for SVP. The NMAEs of SVP on the training data and test data are shown in Fig.~\ref{fig_main_MC_SVP_NMAE_tog}. 
The minimum NMAE for test data is $0.1929$, achieved when the rank is set to $3$.


Lastly, we apply the bi-factored gradient descent (BFGD) algorithm proposed in \cite{park2016finding},\footnote{\url{http://akyrillidis.github.io/projects/}} which is a variant of the projected gradient descent algorithm applied to the non-convex Burer-Monteiro factorization formulation as described in Section~\ref{sec_nonconvex}. We set the maximum number of iterations to $4000$, and the convergence tolerance to $5\times 10^{-6}$. Similarly as before, BFGD typically terminates early in our experiment. For the step size we use the default setting of the above implementation.
\begin{figure}[!htp]
\centering 
\includegraphics[height=1.6in,width=0.4\textwidth]{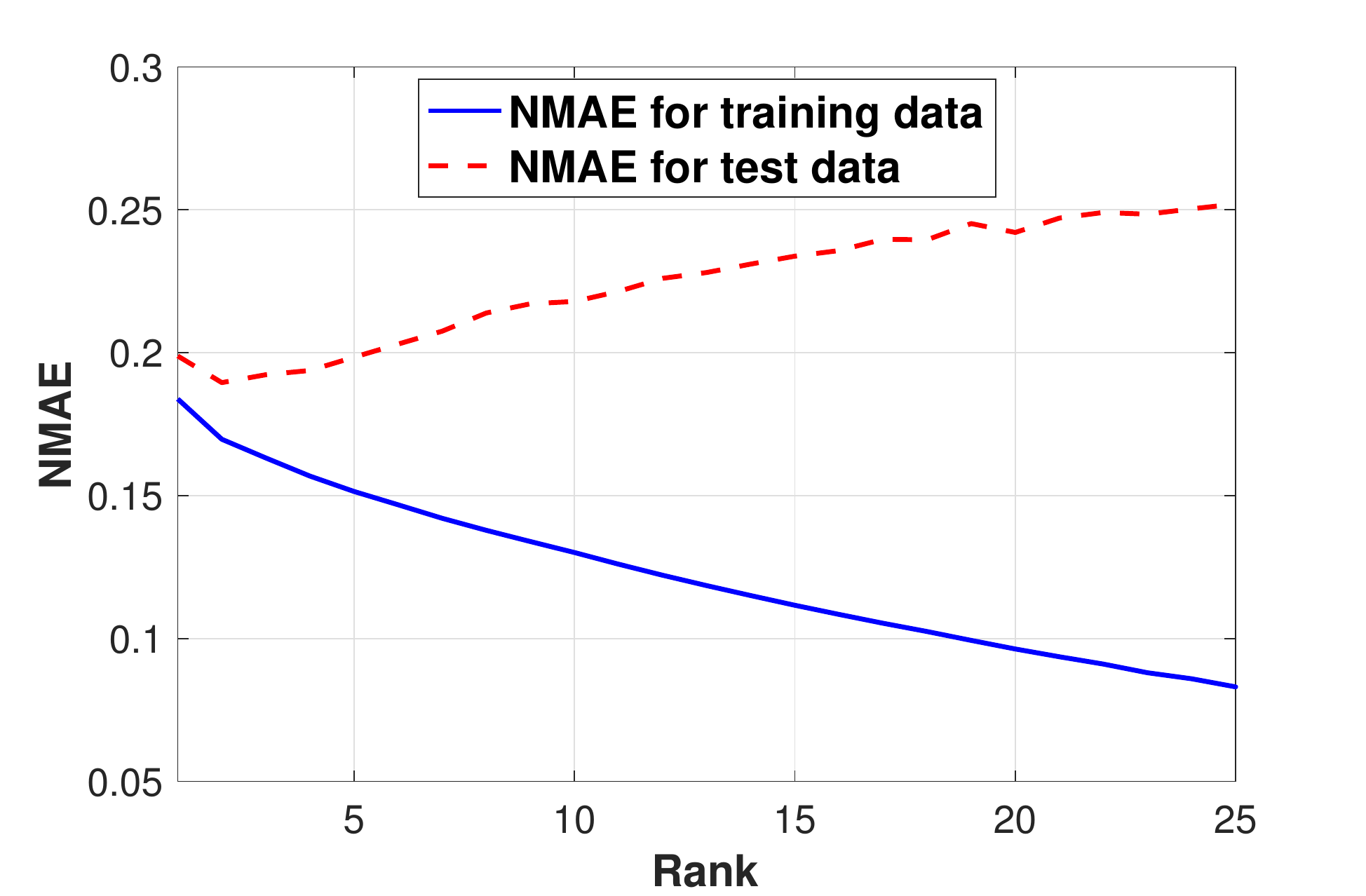}
\caption{Performance of BFGD \cite{park2016finding} on MovieLens 100K dataset: The NMAE for training data and test data via BFGD with respect to the rank.}
\label{fig_main_MC_Factored_Gradient_Descent_I_NMAE_tog}
\end{figure}
The NMAEs of BFGD for the training data and test data  are shown in Fig.~\ref{fig_main_MC_Factored_Gradient_Descent_I_NMAE_tog}. The minimum NMAE for test data is $0.1895$, achieved by setting the rank to $2$.\\

We make several observations from the above experiment results. First, we see that consistently across the three algorithms, the training error generally goes down as the rank becomes larger, whereas the test error exhibits a U-shape behavior, decreasing first and then increasing later. This phenomenon is in accordance with the bias-variance tradeoff principle described in Section~\ref{sec_models}, and in particular shows that using a low-rank model is helpful in reducing the variance and prevents overfitting. Second, all three algorithms achieve a minimum test NMAE around 0.19, using a rank no more than $ 5 $. The small optimal values for the rank are likely due to the highly noisy nature of the MovieLens dataset, for which suppressing variance is crucial to good performance on the test set. Finally, while the estimation/prediction performance of these algorithms is similar, their computational costs, such as running times and memory usage, vary. These costs depend heavily on the specific implementations and termination criteria used, so we do not provide a detailed comparison here.

\section{Concluding Remarks} \label{sec_conclusions}

Low-rank matrices represent an important class of signals with low-dimensional intrinsic structures. In this article, we have presented some recent developments on low-rank matrix estimation, focusing on the setting with incomplete measurements and additional structural constraints. We have particularly emphasized the remarkable modeling power of low-rank matrices, which are useful in a range of problems much wider than the name may suggest, including those where the presence of low-rank structures are not obvious at all. In terms of algorithms and theory, attention is paid to the integration of statistical and computational considerations: fast algorithms have been developed that are applicable to large-scale problems, and at the same time enjoy provable performance guarantees under mild assumptions. As we have seen, such recent progress is made possible by combining techniques from diverse fields; in particular, convex and nonconvex optimization, as well as probabilistic analysis, play a key role. 

We conclude by mentioning a few topics and future directions that are not covered in this article. We have focused on the matrix sensing and completion problems with linear measurements. There are many other low-rank estimation problems that are amenable to convex and nonconvex optimization-based algorithms, and enjoy similar geometric properties and performance guarantees. A partial list of such problems includes phase retrieval~\cite{sun2016geometric}, blind deconvolution \cite{li2016rapid}, robust PCA~\cite{ge2017no}, dictionary learning~\cite{sun2015complete}, lifting for mixture problems~\cite{chen2017mixed_it}, low-rank phase retrieval \cite{vaswani2017low}, community detection~\cite{chen2013planted}, and synchronization problems~\cite{bandeira2016low}. More broadly, applications of low-rank matrix recovery go well beyond the setting of linear measurements and least-squares objectives. Prime examples include low-rank matrix recovery with quantized, categorical and non-Gaussian data~\cite{davenport2014_onebitMC,lafond2015low}, and ranking from comparison-based observations~\cite{lu2015individualized}. These problems involve more general objective functions (such as the log-likelihood) and constraints that depend on the specific observation schemes and noise structures. Another promising line of research aims at exploiting hidden low-rank structures in settings where the problem on the surface has nothing to do with low-rank matrices, yet such structures reveal themselves under suitable transformation and approximation. Problems of this type include latent variable models with certain smoothness/monotonicity properties~\cite{chatterjee2014universal}. 

Another topic of much interest is how to select the model rank automatically and robustly, and how to quantify the effect of model mismatch. These are important issues even in standard matrix sensing and completion; we have not discussed these issues in detail in this survey. Finally, we have omitted many other low-rank recovery algorithms that are not directly based on (continuous) optimization, including various spectral methods, kernel and nearest neighbor type methods, and algorithms with a more combinatorial flavor. Some of these algorithms are particularly useful in problems involving complicated discrete and time-evolving structures and active/adaptive sampling procedures. All of these topics are the subject of active research with tremendous potential.

\section*{Acknowledgment}

The authors thank Mr.\ Yuanxin Li for preparing the numerical experiments in this paper. The work of Y.\ Chen is supported in part by NSF under the CRII award 1657420 and
grant CCF-1704828. The work of Y.\ Chi is supported in part by AFOSR under the grant FA9550-15-1-0205, by ONR under the grant N00014-18-1-2142, and by
NSF under the grants CAREER ECCS-1818571 and CCF-1806154.

 \bibliographystyle{IEEEtran}

\bibliography{lowrank_spm_all}

%

%
%

\end{document}

%% file: chen_macro.tex
\usepackage{color}

\newcommand{\notnow}[1]{}

\makeatother

\newcommand{\defn}{\ensuremath{: \, =  }}
\newcommand{\order}{\ensuremath{\mathcal{O}}}

\newcommand{\real}{\ensuremath{\mathbb{R}}}

\newcommand{\OneMat}{\ensuremath{\boldsymbol{J}}} 

\DeclareMathOperator{\rank}{rank}

\DeclareMathOperator{\poly}{poly}

\newlength{\widebarargwidth}
\newlength{\widebarargheight}
\newlength{\widebarargdepth}

\newcommand{\CROSS}[2]{\ensuremath{#1 {#2}^T }}

\newcommand{\matsnorm}[2]{\Vert #1 \Vert_{{#2}}}
\newcommand{\vecnorm}[2]{\| #1\|_{#2}}

\newcommand{\twonorm}[1]{\vecnorm{#1}{2}}

\newcommand{\opnorm}[1]{\ensuremath{\matsnorm{#1}{}}}

\newcommand{\twoinfnorm}[1]{\ensuremath{\vecnorm{#1}{2,\infty}}}

\newcommand{\usedim}{\ensuremath{n}} 
\newcommand{\rdim}{\ensuremath{r}} 
\newcommand{\numobs}{\ensuremath{m}} 

\newcommand{\noisestd}{\ensuremath{\sigma}} 

\newcommand{\yout}{\ensuremath{ \by }}  
\newcommand{\Yout}{\ensuremath{ \bY }} 

\newcommand{\Loss}{\ensuremath{F}}
\newcommand{\FLoss}{\ensuremath{f}}
\newcommand{\EmpLoss}{\ensuremath{\Loss}}

\newcommand{\EmpFLoss}{\ensuremath{\FLoss}}

\newcommand{\LossTil}{\EmpFLoss}

\newcommand{\ltheta}{\bL}
\newcommand{\rtheta}{\bR}

\newcommand{\PlTheta}{\ensuremath{\bX}}

\newcommand{\lthetait}[1]{\ensuremath{\ltheta^{#1}}}
\newcommand{\rthetait}[1]{\ensuremath{\rtheta^{#1}}}

\newcommand{\stepit}[1]{\ensuremath{\eta^{#1}}}

\newcommand{\LCons}{\ensuremath{\mathcal{L}}}
\newcommand{\RCons}{\ensuremath{\mathcal{R}}}
\newcommand{\FConsit}[1]{\ensuremath{\mathcal{F}^{}}}

\newcommand{\GenConsit}[1]{\ensuremath{\mathcal{F}^{}}}

\newcommand{\Proj}{\ensuremath{\cP}}

\newcommand{\SVD}{\ensuremath{\textsf{SVD}}}

\newcommand{\BM}{\text{Burer-Monteiro}\xspace}

\newcommand{\Xit}[1]{\bA_{#1}} 
\newcommand{\Xmap}{\mathcal{A}}
\newcommand{\ripparam}[1]{\delta_{#1}}

\newcommand{\inco}{\ensuremath{\mu}} 
\newcommand{\pobs}{\ensuremath{p}} 
\newcommand{\Obs}{\ensuremath{\Omega}} 
\newcommand{\ProjObs}{\ensuremath{\Proj_{\Obs}}}

\newcommand{\noisemat}{\ensuremath{ \bW }}  

\newcommand{\csize}{\ensuremath{\ell}}  

\newcommand{\pin}{\ensuremath{\tau_{\text{in}}}}
\newcommand{\qout}{\ensuremath{\tau_{\text{out}}}}

